\documentclass[3p,11pt]{elsarticle}

\usepackage{graphicx}
\usepackage{longtable}
\usepackage{booktabs}
\usepackage{multirow}
\usepackage{array}
\usepackage{calc}
\usepackage{enumitem}
\usepackage{algorithm}
\usepackage{algorithmic}
\usepackage{amsthm}
\usepackage{amsmath}
\usepackage{float}
\usepackage{stfloats}
\usepackage{caption}
\usepackage{placeins}
\usepackage{amssymb}
\usepackage[colorlinks,linkcolor=blue]{hyperref}
\usepackage{lineno}

\journal{}

\newif\ifblind
\blindfalse

\begin{document}

\begin{frontmatter}

\title{Agentic ERP: Multi-Agent Large Language Model Architecture for Autonomous Enterprise Resource Planning}

\ifblind
\author{Anonymous Author(s)}
\else
\author[kth]{Zhihao Liu\corref{cor1}}
\ead{zhihaoliu@ieee.org}
\cortext[cor1]{Corresponding author}

\author[kth]{Tianyu Wang}
\author[kth]{Xi Vincent Wang}
\author[kth]{Lihui Wang}

\affiliation[kth]{organization={Department of Production Engineering, KTH Royal Institute of Technology},
            city={Stockholm},
            country={Sweden}}
\fi

\begin{keyword}
Expert systems \sep Large language model agents \sep Enterprise resource planning \sep Decision support \sep Multi-agent orchestration \sep Human-in-the-loop automation
\end{keyword}

\begin{abstract}
Enterprise Resource Planning (ERP) systems record transactions reliably but still delegate almost all operational decision-making to human specialists, because classical rule-based automation cannot reason about exceptions and monolithic AI assistants degrade when asked to coordinate across functional boundaries. This paper presents \emph{Agentic ERP}, an expert-system architecture that combines role-aligned large-language-model (LLM) agents with a risk-tiered human-in-the-loop harness and a graph-based orchestrator to execute end-to-end business workflows on a production ERP backend. First, autonomous ERP operation is formulated as a constrained sequential-decision problem over a structured enterprise state, with a decomposition argument linking role-aligned agents to a measurable reduction in per-step tool-selection complexity. Second, a graph-based Planner--Executor--Reflector--Responder orchestration decouples generation from evaluation through externalised grading criteria and sprint contracts, packaging recent harness-engineering principles as inspectable expert-system artefacts. Third, the system is evaluated at three levels: a scenario-based task suite, a comprehensive comparison of six orchestration paradigms on cross-functional crisis tasks, and a 365-day agent-in-the-loop simulation against rule-based RPA and no-intervention baselines. Across these levels the proposed multi-agent method is significantly better than the baseline, and the system sustains a simulated year of operation with zero stockouts while the rule-based baseline accumulates hundreds under the same demand stream. The work shows that role-aligned LLM agents under human oversight can move an ERP system from passively recording transactions to actively executing operational decisions, and it provides a reference architecture and an evaluation protocol for autonomous enterprise resource planning.

\end{abstract}

\end{frontmatter}

\section{Introduction}
\label{sec:intro}

Enterprise Resource Planning (ERP) systems have become the operational backbone of modern organizations, integrating business functions including sales, procurement, manufacturing, inventory management, human resources, and finance into unified information architectures \citep{mahapatra2025dynamic}. The global ERP software market reached approximately USD 59 billion in 2023, reflecting the continued strategic importance organizations place on ERP systems for integrating enterprise-wide processes, improving operational efficiency, and supporting faster, more informed decision-making \citep{gartner2025erp, grandview2026erp}.

Despite their ubiquity and strategic importance, ERP implementations face persistent challenges that limit their effectiveness in practice. Most fundamentally, ERP systems remain heavily dependent on manual intervention for decision-making. While these systems excel at recording transactions and enforcing business rules, the cognitive work of interpreting situations, making judgments, and deciding on appropriate actions still falls predominantly to human operators \citep{shaul2013critical}. A purchasing officer must still decide when to reorder, from which supplier, and in what quantity; an accountant must still review anomalies and determine appropriate responses; a sales representative must still evaluate customer requests and negotiate terms.
Where automation does exist within ERP systems, it relies on rigid, rule-based approaches that struggle with exceptions and novel situations. Workflow automation can route documents and trigger notifications based on predefined conditions, but cannot reason about situations that fall outside programmed rules \citep{van2018robotic}. When a supplier delivery is delayed, when a customer requests non-standard terms, or when inventory levels suggest an unusual demand pattern, human judgment is required.
ERP systems also present technical interfaces that exclude non-specialist users. The complexity of ERP navigation, with its hundreds of screens, fields, and transaction codes, creates a barrier that limits who can effectively operate these systems. Organizations invest substantial resources in training, yet many users interact with only a small fraction of available functionality \citep{grabski2011enterprise}. Business stakeholders who need information or want to execute transactions often must work through intermediaries rather than interacting with the system directly.
Coordination across functional boundaries poses a further, persistent challenge. While ERP systems integrate data across departments, the work of coordinating activities across them still requires manual effort: sales commitments must align with inventory availability, production schedules must reflect procurement lead times, and financial implications must be weighed in operational decisions.

Recent large language models (LLMs) exhibit four properties that map onto the four limitations identified above \citep{achiam2023gpt, rieger2026possibilities}. First, they interpret unstructured queries directly, removing the need to traverse interface hierarchies or formulate structured queries: a request such as ``check if we can fulfil the Acme order with current inventory'' can be acted on without intermediate translation. Second, they reason in multiple steps, decomposing requests and adapting to intermediate observations rather than executing fixed sequences, which permits responses to customer complaints, anomalies, or unusual demand patterns to be derived from evidence rather than from a pre-programmed branch. Third, they invoke external tools with appropriate parameters \citep{schick2023toolformer, wang2026llm}, allowing the LLM to mediate between user intent and the typed APIs exposed by an enterprise system. Fourth, they synthesise information across domains without per-combination engineering, so that a question linking production delay to financial exposure can draw on manufacturing, inventory, sales, and accounting evidence in a single response. None of these properties is sufficient on its own to deliver autonomous enterprise operation, but together they mark a qualitative departure from the rule-based and chatbot paradigms that dominate prior enterprise automation.

Previous studies, however, do not amount to a system that can run an enterprise, and three strands of existing work each address a part, but not the whole, of the autonomous-ERP problem. General-purpose autonomous agents such as AutoGPT \citep{yang2023auto}, BabyAGI \citep{talebirad2023multi}, and multi-agent frameworks \citep{wu2023autogen,hong2024metagpt,qian2024chatdev} demonstrate that LLM agents can pursue open-ended goals, but they do not engage with the strong consistency, auditability, and role-separation requirements of production ERP systems. Enterprise AI research has delivered narrow decision-support components such as forecasting \citep{makridakis2018statistical,kraus2020deep}, anomaly detection \citep{chandola2009anomaly}, and dialog systems \citep{xu2017new}, all of which inform humans rather than execute on their behalf. Classical expert systems and decision-support literature \citep{kwon2001multi, liu2011empirical, arnott2014review} offer formal treatments of knowledge representation, human-in-the-loop control, and multi-agent coordination, but pre-date modern LLM reasoning. What is missing is an integrated expert-system design in which LLM reasoning, knowledge-based control, and ERP transactional guarantees coexist, together with an empirical protocol that measures whether such a system can actually run an enterprise.

This paper addresses that gap in terms of three aspects. The first concerns architecture: how an LLM-based expert system should be decomposed across enterprise functions so that tool-selection complexity, inter-domain coordination, and human oversight can all be handled under the transactional guarantees of a production ERP backend. The second concerns orchestration: how a graph-based Planner--Executor--Reflector--Responder pipeline compares against representative LLM-agent paradigms (ReAct, Plan-and-Solve, Function Calling, AutoGen, CrewAI) on enterprise decision scenarios that require cross-functional reasoning. The third concerns sustained operation: how the proposed system behaves over a year-long agent-in-the-loop simulation driven by stochastic demand and supply, relative to rule-based RPA and no-intervention baselines, on both operational-quality and financial outcomes. The contributions of this paper are summarised as below.

\begin{enumerate}[label=(\roman*)]
    \item Formalization of autonomous ERP operation as a constrained sequential-decision problem over a structured enterprise state, with a state-space decomposition result that links role-based agent design to a multiplicative reduction in tool-selection complexity.
    \item Role-aligned multi-agent design comprising five agents (ERP Coordinator, Sales, Inventory, Purchasing, Finance) whose scopes mirror the organisational roles documented in the decision-support literature, together with a risk-tiered human-in-the-loop harness that gates every write against configurable monetary, categorical, and anomaly-based triggers.
    \item Orchestration as a knowledge artefact: a Planner--Executor--Reflector--Responder graph that decouples generation from evaluation and externalises grading criteria and sprint contracts as inspectable objects, operationalising recent harness-engineering principles in an expert-system setting.
    \item A quantitative evaluation protocol with three complementary layers: (a) a scenario-based task suite across six functional categories, (b) a controlled head-to-head comparison of six orchestration paradigms on high-stakes cross-functional crisis scenarios, and (c) a 365-day agent-in-the-loop simulation paired with rule-based RPA and no-intervention baselines.
\end{enumerate}

The paper proceeds as follows. Section~\ref{sec:related} positions the work against the LLM-agent, multi-agent, enterprise-AI, and harness-engineering literature, and states the gap the paper targets. Section~\ref{sec:methodology} formalises the problem, develops the theoretical framework, and describes the architecture, coordination protocols, and orchestration algorithms. Section~\ref{sec:experiments} defines the experimental design, including the scenario suite, orchestration benchmark, long-horizon simulation, and metrics. Section~\ref{sec:results} reports the experimental results. Section~\ref{sec:discussion} discusses the principal findings and limitations. Section~\ref{sec:conclusion} concludes this work.

\section{Related Work}
\label{sec:related}

\subsection{LLM-Based Autonomous Agents}

The paradigm of LLM-based agents emerged following demonstrations that large language models possess emergent capabilities for reasoning, planning, and tool use \citep{brown2020language, wei2022emergent}. Several foundational works established key architectural patterns.
ReAct \citep{yao2022react} introduced the reasoning-action interleaving paradigm, where LLMs alternate between generating thoughts (reasoning traces) and actions (tool invocations). This approach enables models to combine deliberative reasoning with external interactions, improving performance on tasks requiring multi-step problem solving. The ReAct pattern informs our agent design, where each agent reasons about requests before executing ERP operations.
Toolformer \citep{schick2023toolformer} demonstrated that language models can learn to use external tools in a self-supervised manner, determining when tool use is beneficial and generating appropriate API calls. This work established that tool-augmented LLMs can function as practical autonomous systems rather than mere text generators.
Building on these foundations, autonomous agent systems emerged including AutoGPT \citep{yang2023auto} and BabyAGI \citep{talebirad2023multi}, which demonstrated LLMs operating with increasing autonomy to accomplish complex goals. However, these systems revealed challenges in reliability, controllability, and alignment that must be addressed for enterprise deployment.
More recent work has focused on improving agent reliability. Reflexion \citep{shinn2023reflexion} introduced self-reflection mechanisms enabling agents to learn from failures. Tree of Thoughts \citep{yao2023tree} improved planning through structured exploration of solution spaces. These advances improve the robustness required for enterprise applications where errors have operational consequences.
Despite these advances, existing agent frameworks remain fundamentally designed for short-horizon, single-session tasks, and thus cannot directly accommodate ERP workflows that demand persistent state, cross-departmental coordination, and auditable execution over extended timescales.

\subsection{Multi-Agent LLM Systems}

Recognizing that complex tasks often exceed single-agent capabilities, researchers have developed frameworks for multi-agent coordination with LLMs and their applications \citep{sadak2026multi, wu2026intention, li2026penexpert}.
CAMEL \citep{li2023camel} explored role-playing communication between LLM agents, demonstrating that agents assigned distinct roles can collaborate through structured dialogue. While focused on general conversation, this work established the viability of role-based agent specialization.
MetaGPT \citep{hong2024metagpt} applied multi-agent LLM systems to software development, assigning agents roles such as Product Manager, Architect, and Engineer. The system demonstrated that mimicking human organizational structures can improve output quality and coordination. This finding directly informs our decision to align agents with established ERP operator roles.
AutoGen \citep{wu2023autogen} provides a framework for conversational multi-agent systems, supporting various interaction patterns including hierarchical orchestration and peer-to-peer dialogue. The framework's flexibility influenced our coordination protocol design.
LangGraph \citep{langgraph2026} introduced graph-based abstractions for stateful agent coordination, enabling complex workflows with conditional branching, cycles, and persistent state. We adopt LangGraph as our orchestration foundation, leveraging its capabilities for managing multi-agent enterprise workflows.
ChatDev \citep{qian2024chatdev} demonstrated end-to-end software development through multi-agent collaboration, highlighting both the potential and challenges of coordinated LLM agents on complex tasks. Their findings on the importance of clear role definitions and communication protocols inform our agent design.
While these frameworks demonstrate the maturity of multi-agent coordination mechanisms, they have been validated primarily in domains such as software development and open-ended dialogue, where errors are recoverable and outputs are advisory. Agentic ERP, by contrast, requires agents to execute binding business transactions against a system of record, under organizational constraints including role-based access control, approval hierarchies, and audit obligations—demands that existing frameworks leave unaddressed.

\subsection{AI Applications in Enterprise Systems}

Traditional AI applications in enterprise contexts have focused on specific, well-defined problems rather than general autonomous operation.
Demand forecasting represents one of the most established applications, with techniques ranging from statistical methods to deep learning approaches applied to predict future demand and optimize inventory levels \citep{syntetos2016supply, makridakis2018statistical, chu2026human}. However, these systems typically provide predictions that humans must interpret and act upon, rather than executing decisions autonomously.
Anomaly detection systems identify unusual patterns in transactions, operations, or behaviors that may indicate errors, fraud, or operational issues \citep{chandola2009anomaly}. While valuable for directing human attention, these systems similarly require human judgment to determine appropriate responses.
Chatbot interfaces have been deployed in enterprise contexts for customer service and employee self-service, handling routine inquiries and transactions \citep{xu2017new}. However, these systems typically operate with narrow, pre-defined capabilities rather than general enterprise operation authority.
Robotic Process Automation (RPA) automates repetitive tasks through UI-level scripting, mimicking human interactions with enterprise applications \citep{van2018robotic}. RPA has achieved significant adoption for high-volume, rule-based processes but lacks the reasoning capability to handle exceptions or adapt to changing conditions. Our experimental comparison against an RPA baseline quantifies the benefits of LLM-based reasoning over rigid automation.
Recent work has explored LLMs in enterprise contexts for document processing \citep{koh2022empirical}, knowledge management \citep{lewis2020retrieval}, and code generation \citep{chen2021evaluating}. However, applications to autonomous operational execution, where AI systems make and execute business decisions with real consequences, remain largely unexplored.

\subsection{Harness Engineering for Long-Running Agents}

Unlike conversational question answering, ERP workflows are inherently long-horizon: a single business process (e.g., procure-to-pay) traverses multiple steps and departments, with execution timelines that far exceed the lifetime of an individual LLM session. Consequently, conventional one-shot agent patterns are inadequate, and an execution harness is required to manage persistent state, orchestrate subtasks, and resume execution across session boundaries.

Recent harness-engineering work addresses the failure modes of LLM agents that operate over extended task horizons and multiple context windows \citep{anthropic2025harness,anthropic2026harness}. Several observations from that literature bear directly on the present architecture. Long-running agents exhibit ``context anxiety,'' terminating prematurely as the perceived context budget approaches its limit, and structured handoffs that reset context while preserving the necessary state mitigate this more effectively than simple compaction. Agents that evaluate their own work display a systematic positive bias and approve mediocre outputs, a problem analogous to discriminator collapse in generative adversarial networks, which a separately configured evaluator agent calibrated with explicit grading criteria can counter by supplying an external signal. An explicit acceptance contract negotiated between planner and evaluator before execution gives a testable definition of completion that prevents both early termination and scope creep, while incremental progress with auditable checkpoints enables recovery across context boundaries that would otherwise truncate execution. These mechanisms are important precisely because they determine whether a long-running agent stays reliable, and they motivate treating grading criteria, sprint contracts, and reflection loops as first-class, inspectable artefacts rather than implementation details.

\subsection{Expert Systems, Decision Support, and the Role of Humans}

The contemporary LLM-agent literature recapitulates, in a new substrate, several questions first posed in the tradition of expert systems and decision support systems (DSS) \citep{arnott2014review}. That tradition identified four properties that any production knowledge-based system must provide: (a) an explicit and inspectable knowledge representation; (b) a reasoning component whose steps can be traced; (c) an integration path with transactional systems that enforce business rules; and (d) a deliberate policy for where human judgement enters the decision loop. Modern multi-agent frameworks excel at (b) but provide only implicit mechanisms for (a), (c), and (d). Work on intelligent agents for organisations formalised role-based decomposition as an organisational-design choice. More recent human-in-the-loop machine-learning surveys \citep{mosqueira2023human} emphasise that oversight granularity must be designed together with the decision process, not bolted on. Our architecture takes these four properties as first-class design requirements: skills and grading criteria are externalised as files, the orchestrator emits an inspectable plan, the ERP backend retains authority over writes, and every high-risk action is gated by a configurable approval policy.

\subsection{Research Gap}

Existing work has demonstrated LLM tool use, multi-agent coordination, enterprise decision support, and human-in-the-loop control separately. What remains missing is an auditable, role-separated, transaction-safe architecture that lets LLM agents execute ERP operations under explicit oversight, together with an evaluation protocol beyond isolated tasks.  General-purpose LLM-agent work \citep{yao2022react,schick2023toolformer,wu2023autogen,hong2024metagpt,wang2024survey,xi2025rise} demonstrates reasoning and tool use on open-ended or code-centric tasks but does not engage with ERP-specific transactional, auditability, and role-separation constraints. Enterprise AI work \citep{syntetos2016supply,chandola2009anomaly,xu2017new,kraus2020deep, jiang2025towards, wang2026intelligent} yields strong point solutions that still require a human to take the resulting action. Classical expert-system and DSS research \citep{arnott2014review} supplies the right vocabulary for knowledge representation, role decomposition, and human oversight, but pre-dates contemporary LLM reasoning and tool use. Harness-engineering guidance \citep{anthropic2025harness,anthropic2026harness} identifies the failure modes of long-running agents but does not operationalise them for production ERP. No existing work integrates these four strands into a single architecture and evaluates it under both scenario-level and year-long agent-in-the-loop conditions. The remainder of this paper presents that integration.

\section{Methodology}
\label{sec:methodology}

\subsection{Problem Formulation}

We formalise autonomous enterprise operation as a constrained partially-observable sequential decision problem over a structured enterprise state, with multiple specialised decision-makers whose observations are restricted to disjoint subspaces and whose actions are mediated by typed tool APIs. Table~\ref{tab:notation} summarises the notation used throughout this section.

\begin{table}[H]
\centering
\caption{List of Notation.}
\label{tab:notation}
\small
\begin{tabular}{p{4cm}p{10cm}}
\toprule
\textbf{Symbol} & \textbf{Definition} \\
\midrule
$\mathbb{S}$ & space of all enterprise states; a particular state is $\mathcal{S} \in \mathbb{S}$ \\
$\mathcal{S}_t$ & enterprise state at logical time $t$: the tuple $\langle S_{\text{inv}}, S_{\text{ord}}, S_{\text{fin}}, S_{\text{sup}}, S_{\text{cust}}\rangle$ of inventory, order, financial, supplier, and customer sub-states (Eq.~\ref{eq:state}) \\
$\mathcal{Q}$ & set of admissible natural-language queries, each expressing one business intent \\
$\mathcal{A} = \{\alpha_i\}_{i=1}^{n}$ & population of the $n$ role-aligned agents; agent $\alpha_i$ is defined by its domain scope $D_i$, tool set $T_i$, and policy $\pi_i$ \\
$\mathcal{T} = \bigcup_{i=1}^{n} T_i$ & global tool space, the union of all agents' tool sets; $|\mathcal{T}| = 46$ in our implementation \\
$\bar{k}$ & mean per-agent tool-set size, $\bar{k} = \mathbb{E}_i[|T_i|]$ ($\bar{k} = 9.2$ in our implementation) \\
$\Omega$ & orchestration function composing the Plan, Execute, Reflect, and Respond stages (Eq.~\ref{eq:orchestration}) \\
$\mathcal{G} = \{(d_i, w_i, \theta_i)\}$ & grading rubric whose triples each give a quality dimension $d_i$, its weight $w_i$, and its acceptance threshold $\theta_i$ (Section~\ref{sec:orchestration_algo}) \\
$\mathcal{C} = (\mathbf{a}, \mathbf{d}, t_0)$ & sprint contract: acceptance criteria $\mathbf{a}$, expected deliverables $\mathbf{d}$, and creation time $t_0$ \\
$\epsilon_r$ & routing error, the probability that the router assigns a query to the wrong agent \\
$\epsilon_{\text{sel}}(k)$ & per-step tool-selection error when an agent chooses among $k$ candidate tools \\
$c_{\text{op}}, c_{\text{api}}, c_{\text{err}}$ & operational-cost, API-cost, and error-penalty functionals in the objective (Eq.~\ref{eq:objective}) \\
$\lambda_1, \lambda_2$ & non-negative weights trading API cost and error penalty against operational cost (Eq.~\ref{eq:objective}) \\
$H$ & horizon length: the number of queries in the stream $\{q_t\}_{t=1}^{H}$ \\
$K$ & maximum number of reflect--replan iterations permitted per query \\
$p$ & probability that a single execution attempt yields an output satisfying $\mathcal{G}$ \\
\bottomrule
\end{tabular}
\end{table}

The object the system operates on is the \emph{enterprise state}. At any logical time it is the tuple
\begin{equation}
\label{eq:state}
\mathcal{S} = \langle S_{\text{inv}}, S_{\text{ord}}, S_{\text{fin}}, S_{\text{sup}}, S_{\text{cust}} \rangle ,
\end{equation}
whose five components capture the operational sub-domains of the business. Inventory $S_{\text{inv}} = \{(p_i, x_i, l_i)\}_{i=1}^{n_p}$ records each product $p_i$ with its on-hand quantity $x_i$ and location $l_i$; the order book $S_{\text{ord}} = \{(o_j, c_j, \tau_j, \sigma_j)\}_{j=1}^{n_o}$ records each order $o_j$ with its customer $c_j$, timestamp $\tau_j$, and status $\sigma_j$; the financial state $S_{\text{fin}} = (B, R, P, \mathbf{r}, \mathbf{p})$ holds the bank balance $B$, receivables $R$, payables $P$, and the receivable and payable aging vectors $\mathbf{r}$ and $\mathbf{p}$; the supplier state $S_{\text{sup}} = \{(s_k, \ell_k, \rho_k)\}_{k=1}^{n_s}$ holds each supplier $s_k$ with lead time $\ell_k$ and reliability $\rho_k$; and the customer state $S_{\text{cust}} = \{(c_m, h_m, \kappa_m)\}_{m=1}^{n_c}$ holds each customer $c_m$ with purchase history $h_m$ and credit limit $\kappa_m$. The space of all such states is written $\mathbb{S}$. This partition is not incidental: it is the same boundary along which agents are later assigned, so that each agent observes only the sub-state relevant to its role.

The system is driven by natural-language queries. A query $q \in \mathcal{Q}$ is a free-text utterance expressing a business intent, and a semantic mapping $\phi: \mathcal{Q} \to \mathcal{I}$ classifies it into one of three intent types, $\mathcal{I} = \{i_{\text{info}}, i_{\text{action}}, i_{\text{decision}}\}$, corresponding to an information request, a state-changing action, or a decision that requires judgement; the orchestrator handles each differently.

Operations on the state are carried out by a population of agents $\mathcal{A} = \{\alpha_1, \ldots, \alpha_n\}$. Each agent $\alpha_i$ is characterised by a domain scope $D_i$, the projection of the state space $\mathbb{S}$ onto the components relevant to its role; a tool set $T_i = \{t^i_1, \ldots, t^i_{k_i}\}$ of typed operations with explicit preconditions and effects; and a policy $\pi_i: D_i \times \mathcal{Q} \to \Delta(T_i)$ mapping the observed sub-state and the query to a distribution over tool invocations. The tools of all agents form the global action space $\mathcal{T} = \bigcup_{i=1}^{n} T_i$, with $|\mathcal{T}| = 46$ in our implementation. Each agent exposes only its own $T_i$ rather than the full $\mathcal{T}$, and the consequences of this choice are analysed in the next subsection.

A single query is resolved by an orchestration function $\Omega: \mathcal{Q} \times \mathbb{S} \to \mathcal{A}^* \times \mathcal{T}^*$ that maps a query and the current state to a sequence of agent activations and tool executions. A single pass of $\Omega$ composes four stages,
\begin{equation}
\label{eq:orchestration}
\Omega(q, \mathcal{S}) = \text{Respond} \circ \text{Reflect} \circ \text{Execute} \circ \text{Plan}(q, \mathcal{S}) ,
\end{equation}
in which the planner decomposes the query into sub-tasks, the executor invokes the relevant agent tools, the reflector checks the result against the grading criteria, and the responder synthesises the final answer. The reflector may trigger a bounded replan, so a full run repeats this pass at most $K$ times before returning; the stages and the loop are detailed in Section~\ref{sec:orchestration_algo}.

The objective is not to resolve a single query but to keep the business healthy across a stream of them. Given an initial state $\mathcal{S}_0$ and a query stream $\{q_t\}_{t=1}^{H}$ over a horizon of $H$ steps, let $a_t = \Omega(q_t, \mathcal{S}_t)$ be the action taken at step $t$, after which the state evolves from $\mathcal{S}_t$ to $\mathcal{S}_{t+1}$ under that action and exogenous events. We seek an orchestration policy that minimises the expected cumulative operational cost,
\begin{equation}
\label{eq:objective}
\min_{\Omega} \; \sum_{t=1}^{H} \mathbb{E}\!\left[ c_{\text{op}}(\mathcal{S}_t, a_t) + \lambda_1\, c_{\text{api}}(a_t) + \lambda_2\, c_{\text{err}}(a_t) \right],
\end{equation}
subject to business constraints that must hold at every step $t$: inventory at or above safety stock ($x_i \geq x_i^{\min}$ for every product $i$), liquidity above a floor ($B \geq B^{\min}$), customer credit within limit ($R_c \leq \kappa_c$ for every customer $c$, with $R_c$ the receivable outstanding for $c$), and response latency bounded ($\tau_{\text{response}} \leq \tau^{\max}$). Here $c_{\text{op}}$ penalises operational failures such as stockouts and delays, $c_{\text{api}}$ is the LLM API cost, $c_{\text{err}}$ is an error penalty, and the non-negative weights $\lambda_1, \lambda_2$ trade the three terms off. We do not solve~\eqref{eq:objective} in closed form; it states the target that the architecture of the following sections is designed to pursue, and against which the evaluation in Section~\ref{sec:results} measures the system.

\subsection{Design Principles}
\label{sec:principles}

The architecture rests on three design principles, each supported by a structural argument and stated as a testable prediction that the ablation study evaluates. The arguments are not a substitute for measurement; they motivate the architectural choices and bound the quantities the experiments later estimate. The three principles are (a) narrow per-agent scope, so that tool selection remains tractable; (b) separation of generation from evaluation, so that errors are detected rather than propagated; and (c) parallel coordination wherever data dependencies allow.

The first principle, role-based decomposition, lowers tool-selection error under a mild assumption on how that error scales. Partition the global tool space $\mathcal{T}$ into the disjoint per-agent sets $T_i$, $i = 1, \ldots, n$, with mean size $\bar{k} := \mathbb{E}_i[|T_i|]$, and suppose a router $r: \mathcal{Q} \to \mathcal{A}$ identifies the relevant agent with accuracy $1 - \epsilon_r$. The argument relies on the empirical regularity that an LLM's per-step tool-selection error grows with the number of candidate tools, $\epsilon_{\text{sel}}(k_1) \leq \epsilon_{\text{sel}}(k_2)$ whenever $k_1 \leq k_2$, which is a standard property of in-context classification over many labels. Under role-based decomposition a query is served incorrectly only if the router selects the wrong agent, with probability $\epsilon_r$, or the chosen agent $i$ then mis-selects a tool from its own set $T_i$, with probability at most $\epsilon_{\text{sel}}(|T_i|)$. Decomposing on whether the router succeeds, and averaging over which agent it selects, gives
\begin{equation}
\label{eq:decomp}
\epsilon_{\text{MA}} \;\leq\; \epsilon_r + (1 - \epsilon_r)\,\mathbb{E}_i\!\big[\epsilon_{\text{sel}}(|T_i|)\big] ,
\end{equation}
whereas a single agent exposed to every tool incurs $\epsilon_{\text{SA}} = \epsilon_{\text{sel}}(|\mathcal{T}|)$. Because every agent's set is far smaller than the whole, $|T_i| \ll |\mathcal{T}|$, monotonicity makes each term of the average at most $\epsilon_{\text{sel}}(|\mathcal{T}|)$; we write the average compactly as $\epsilon_{\text{sel}}(\bar{k})$, the selection error at the mean set size $\bar{k}$, which equals $\mathbb{E}_i[\epsilon_{\text{sel}}(|T_i|)]$ exactly when $\epsilon_{\text{sel}}$ is affine over this range and closely otherwise. Decomposition therefore improves accuracy whenever $\epsilon_r + (1 - \epsilon_r)\epsilon_{\text{sel}}(\bar{k}) < \epsilon_{\text{sel}}(|\mathcal{T}|)$, that is, when routing is reliable and each agent's tool set is much smaller than the whole. The bound depends on the routing error $\epsilon_r$, which is treated as a quantity to be estimated rather than assumed: given a labelled set with ground-truth routings $\{\alpha^*(q_\ell)\}_{\ell=1}^{L}$, it is estimated as $\widehat{\epsilon}_r = L^{-1}\sum_{\ell} \mathbf{1}[r(q_\ell) \neq \alpha^*(q_\ell)]$ with an accompanying confidence interval. Whether the realised errors satisfy~\eqref{eq:decomp} in our implementation is examined in Section~\ref{sec:results}.

The second principle, separation of generation from evaluation, is realised by a reflection loop in which each candidate output is scored against an explicit grading rubric $\mathcal{G}$, defined alongside the orchestration algorithms in Section~\ref{sec:orchestration_algo}. Its value is clearest by contrast with blind retrying. Let $p \in (0,1]$ be the probability that a single attempt yields an output satisfying $\mathcal{G}$. Were $K$ attempts statistically independent, at least one would succeed with probability
\begin{equation}
\label{eq:reflbound}
1 - (1-p)^K ,
\end{equation}
which rises quickly with $K$. Independence, however, is precisely what retrying without feedback does not provide: an agent that re-attempts a task with no information about why it failed tends to reproduce the same error, so its failures are positively correlated and the realised success rate stays close to the single-attempt value $p$ rather than approaching the independent baseline~\eqref{eq:reflbound}. The reflection loop is designed to close this gap. Because the reflector returns the reason an attempt failed, such as a tool error, a missing field, or an unmet acceptance criterion, the subsequent replan is conditioned on that reason and is less likely to repeat it, so the per-attempt failure probability falls below $1-p$ and the realised success rate moves up towards, and can exceed, the independent baseline. The benefit therefore tracks the information content of reflector feedback rather than the iteration budget $K$, a prediction the ablation in Section~\ref{sec:results} evaluates.

The third principle, parallel coordination, concerns the latency of resolving a single query with several agents, and the design rule is to run agents concurrently whenever their data dependencies allow. Two quantities bound the cost. Running the $m$ agents \emph{serially}, each feeding the next, makes their latencies add and charges a fixed overhead $\tau_c$ for each of the $m-1$ hand-offs, for an expected latency of $\sum_{i} \mathbb{E}[\tau_i] + (m-1)\tau_c$. Running the independent agents \emph{in parallel} instead lets them overlap, so the agents contribute only the latency of the slowest one, $\max_i \mathbb{E}[\tau_i]$, and the only additional cost is a single step that merges their outputs, the aggregation overhead $\tau_{\text{agg}}$, giving an expected latency of $\max_i \mathbb{E}[\tau_i] + \tau_{\text{agg}}$. The two coordination terms are not the same quantity: $(m-1)\tau_c$ is the cumulative cost of the hand-offs a serial schedule performs, whereas $\tau_{\text{agg}}$ is the one-time cost of the merge a parallel schedule performs. Parallelisation lowers latency exactly when the parallel cost is the smaller of the two, which rearranges into a bound on the aggregation overhead,
\begin{equation}
\label{eq:coord}
\tau_{\text{agg}} \;<\; \Big(\sum_{i} \mathbb{E}[\tau_i] - \max_i \mathbb{E}[\tau_i]\Big) + (m-1)\,\tau_c , \qquad m \geq 2 .
\end{equation}
The right-hand side is the latency a serial schedule would spend off the critical path, namely the summed latency of every agent except the slowest, together with the hand-off overhead a parallel schedule avoids. Concurrency is therefore worthwhile as long as merging the agents' outputs costs less than that; for multi-domain queries with several comparably sized agents and a light merge step, the inequality holds with room to spare.

Taken together, these arguments give three falsifiable predictions that the ablation study (Section~\ref{sec:ablation}) evaluates: (P1) single-agent tool-selection error exceeds multi-agent error by a margin that grows with $|\mathcal{T}|$; (P2) the benefit of the reflection loop tracks the information content of reflector feedback rather than the iteration budget $K$ alone; and (P3) parallel coordination reduces latency on multi-domain queries when the aggregation overhead $\tau_{\text{agg}}$ is small relative to the off-critical-path latency it removes.

\subsection{System Architecture Overview}

\begin{figure}[H]
    \centering
    \includegraphics[width=\textwidth]{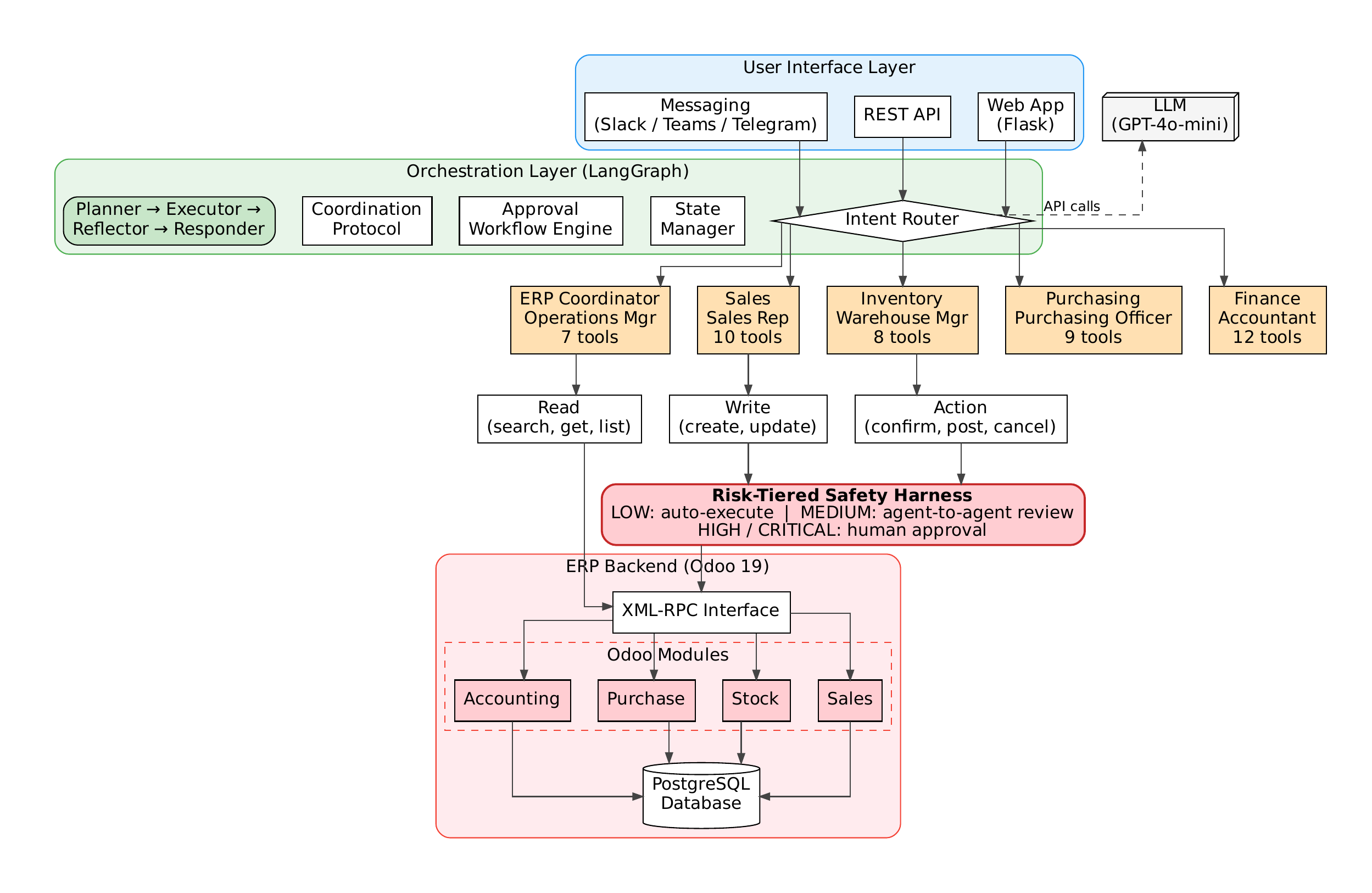}
    \caption{Agentic ERP system architecture. The four-layer design separates concerns: User Interface handles interaction modalities, Orchestration manages routing and coordination, the Agent Layer provides specialized domain expertise, and the ERP Backend ensures data integrity and business rule enforcement.}
    \label{fig:architecture}
\end{figure}

Agentic ERP comprises four architectural layers designed to integrate LLM reasoning capabilities with production ERP systems while maintaining transactional integrity, auditability, and appropriate human oversight. Figure~\ref{fig:architecture} illustrates the overall architecture.

\subsubsection{User Interface Layer}

The user-interface layer exposes three access modalities that share a common request-processing pipeline. A web application provides conversational access with persistent history, allowing users to issue natural-language queries, inspect proposed actions, and approve or modify recommendations before execution. A REST (Representational State Transfer) API exposes the same capabilities programmatically, so that the system can be invoked from external workflows, scheduled tasks, and integration middleware. Connectors for enterprise messaging platforms (Slack, Microsoft Teams, Telegram) extend access to mobile users and to scenarios where the system must react quickly to externally driven events. All three modalities terminate at the same orchestrator entry point, ensuring that observed behaviour does not depend on the access channel.

\subsubsection{Orchestration Layer}

The orchestration layer, implemented in LangGraph, comprises four cooperating components. The intent router analyses incoming requests, identifies the relevant agents through LLM-based classification over agent role descriptions, and, for requests that span multiple domains, selects an appropriate coordination pattern. The state manager persists conversation context, workflow state, and cross-request memory, supporting multi-turn interactions in which subsequent requests build on prior context and long-running workflows that span more than one user session. The approval-workflow engine enforces human-in-the-loop oversight for high-stakes decisions, with configurable policies that route actions to appropriate reviewers according to monetary thresholds, risk categories, and operation types. Finally, a coordination-protocol module manages multi-agent collaboration for complex requests through the sequential, parallel, and hierarchical patterns described in Section~\ref{sec:coordination}.

\begin{figure}[H]
    \centering
    \includegraphics[width=\textwidth]{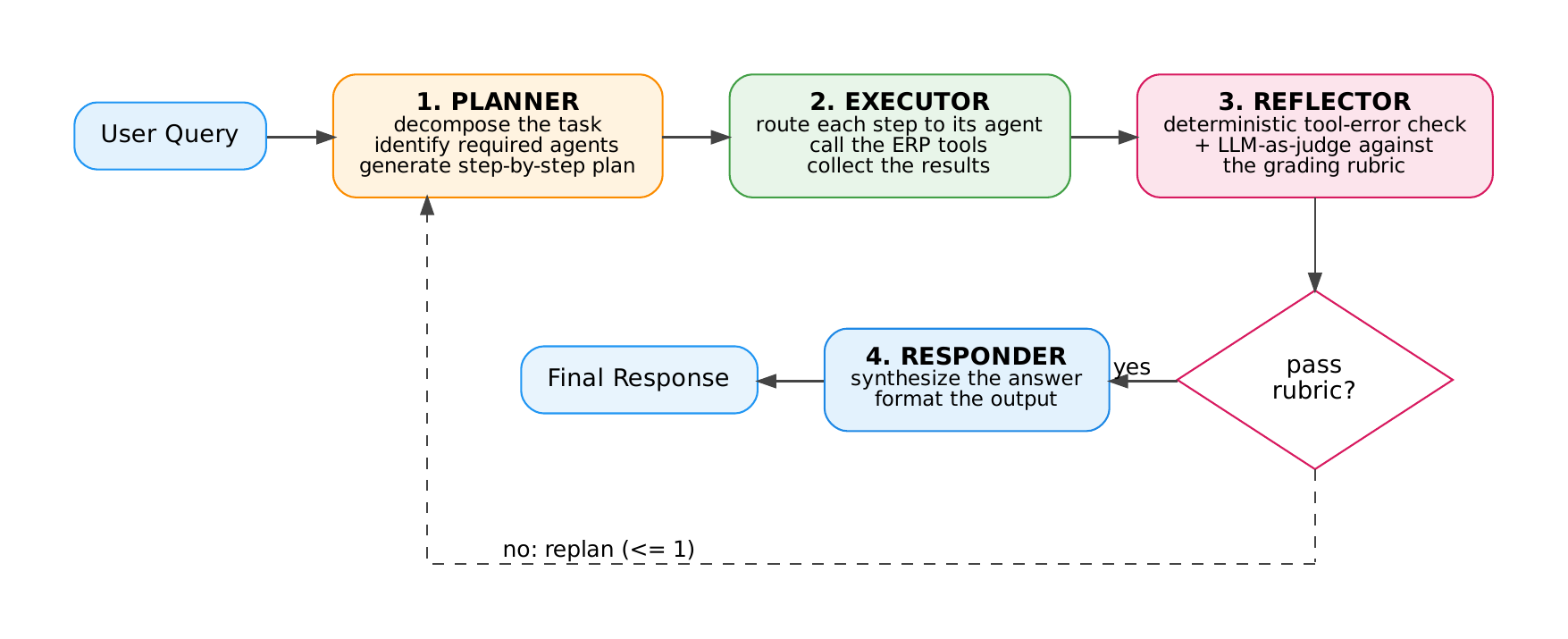}
    \caption{Orchestration workflow. The Planner decomposes the request into a typed-tool plan; the Executor invokes the relevant agent tools; the Reflector applies a deterministic check on tool exceptions plus an LLM-as-judge against grading criteria, and triggers at most one replan on detected failure; the Responder synthesises the final answer.}
    \label{fig:langgraph_workflow}
\end{figure}

Figure~\ref{fig:langgraph_workflow} illustrates the orchestration workflow, showing how queries flow through the Planner--Executor--Reflector--Responder pipeline with self-correction capability. Decoupling generation from reflection is the mechanism intended to catch errors that single-step reasoning would otherwise propagate, evaluated on the crisis benchmark in Section~\ref{sec:e2_scaled}.

\subsubsection{Agent Layer}

The Agent Layer comprises five specialised agents, each aligned with a distinct ERP operator role: an ERP Coordinator (operations manager), Sales, Inventory, Purchasing, and Finance. Choosing the \emph{operator role} as the unit of decomposition, rather than a technical module or an individual tool, follows directly from the decomposition principle of Section~\ref{sec:principles}. Confining an agent to one role keeps its candidate tool set small, an average of $\bar{k} = 9.2$ tools instead of the full $46$, which is the regime in which bound~\eqref{eq:decomp} predicts lower tool-selection error. Role alignment also makes behaviour legible: a Sales Agent that confirms an order acts like the human sales representative a stakeholder already understands, which simplifies debugging and explanation. And it makes the system modular, since a tool or an entire agent can be added or revised without disturbing the others. Table~\ref{tab:agents_detailed} lists the five agents, their roles, and representative tools.

\begin{table}[H]
\centering
\caption{Detailed agent specifications with representative tools}
\label{tab:agents_detailed}
\begin{tabular}{p{3.2cm}p{3.2cm}p{1.2cm}p{6cm}}
\toprule
\textbf{Agent} & \textbf{ERP Role} & \textbf{Tools} & \textbf{Representative Capabilities} \\
\midrule
ERP Coordinator & Operations Manager & 7 & Company dashboard, business health diagnostics, KPI monitoring, cross-department coordination, alert management \\
\addlinespace
Sales & Sales Representative & 10 & Search/create orders, manage customers, generate quotations, check pricing, track top products and customers \\
\addlinespace
Inventory & Warehouse Manager & 8 & Check stock levels, manage locations, process transfers, low stock alerts, inventory valuation \\
\addlinespace
Purchasing & Purchasing Officer & 9 & Request quotations, compare suppliers, create purchase orders, track deliveries, reorder suggestions \\
\addlinespace
Finance & Accountant & 12 & Create invoices, process payments, manage receivables/payables, overdue tracking, financial reporting \\
\midrule
\textbf{Total} & & \textbf{46} & \\
\bottomrule
\end{tabular}
\end{table}

Every agent is constructed the same way, which is what keeps the cost of extension low. A role-specific system prompt fixes the agent's identity, responsibilities, and the boundaries of its authority. Its tools are registered through a uniform interface that associates each tool name with the operation implementing it, and every tool returns a structured result rather than free text: a status flag indicating whether the operation succeeded, together with the returned data or, on failure, an error description. Because outcomes follow this fixed form, the orchestrator can branch on them without interpreting natural language. All five agents share one thin XML-RPC client to Odoo (see section \ref{erp_backend_layer}) rather than re-implementing backend access, so that authentication, transactions, and access control are enforced in exactly one place. Concretely, the scope $D_i$ defined in Section~\ref{sec:methodology} is realised as the set of Odoo models the agent's tools may read and write. Under this construction, adding a capability reduces to registering a tool and adding a role reduces to instantiating an agent, with no change to the orchestrator or the other agents.

\subsubsection{ERP Backend Layer}
\label{erp_backend_layer}

The backend of the system is a production ERP rather than a bespoke datastore, so that the agents operate against the same transactional guarantees and business rules that govern a human operator. We use Odoo~19, which is a widely deployed open-source ERP suite: a modular business-management platform, written in Python over a PostgreSQL database, whose functionality is divided into installable modules that each cover one business domain and define their own data models and operations. The deployment used here installs only the four modules the agents exercise (sales, purchasing, inventory, and accounting) and leaves the many further modules Odoo offers uninstalled, so that the backend contains exactly the sub-systems the trading-company scenario requires.

Odoo was chosen for four reasons. Its open-source licence grants full integration access, including the ability to inspect and, where necessary, modify the internal data models, which a closed commercial ERP would not permit. Its modular architecture lets the relevant sub-systems be deployed in isolation, keeping the backend small and the experiments reproducible. Its development community is active enough to keep the platform and its dependencies current. And it has a documented record of production use in real businesses, which makes results obtained against it more readily transferable to a deployment than results obtained against a simplified backend.

The agents reach Odoo through its XML-RPC interface, the same programmatic interface that third-party software uses to integrate with Odoo. It exposes four capabilities on which the tool layer is built:
\begin{itemize}
    \item \emph{Record manipulation on every data model.} The interface can create, read, update, and delete records of any model in the system (a sales order, a stock movement, a customer, an invoice) which is what lets a single uniform tool pattern cover the entire ERP instead of a bespoke connector per entity.
    \item \emph{Business-method execution with workflow support.} Beyond manipulating records directly, the interface invokes Odoo's domain methods, which encode the platform's business logic and state transitions, such as confirming a quotation into a sales order, validating a delivery, or posting an invoice. Calling these methods triggers exactly the validations and side effects Odoo would apply for a human user, so the agents inherit the ERP's workflow correctness rather than re-implementing it.
    \item \emph{Transactional execution.} Each call runs inside a database transaction, so an operation that fails is rolled back atomically and never leaves a partial write behind; this is the property that lets the safety harness and the tool layer treat every write as all-or-nothing.
    \item \emph{Permission-aware access control.} Every call authenticates as a specific Odoo user and is subject to that user's access rights and record-level rules, so an agent cannot act beyond the permissions granted to its credentials. This gives a second, backend-enforced boundary on top of the role scopes defined in the agent layer.
\end{itemize}

Because these guarantees are enforced by the ERP itself rather than by the agent layer, the backend ensures that, whatever the AI layer infers or attempts, fundamental data integrity and business-rule constraints are preserved.

\subsection{Multi-Agent Coordination}
\label{sec:coordination}

Effective enterprise operation requires coordinating several agents on a single query, which raises two questions: which agents should participate, and how their work should be combined. The first is answered by the two-stage router of Algorithm~\ref{alg:routing}, which returns the active set $\mathcal{A}^*(q) = \{\alpha_i : P(\text{relevant} \mid \alpha_i, q) > \theta\}$ of agents whose estimated relevance exceeds a threshold $\theta$; let $m = |\mathcal{A}^*(q)|$ denote the number of agents it selects. How those agents are combined is governed not by query type but by the data dependencies among their subtasks. Three coordination patterns of increasing generality are distinguished and summarised in Table~\ref{tab:coordination}, complemented by a policy for resolving conflicts between agents.

\emph{Sequential coordination} is appropriate when the subtasks form a dependency chain, each consuming the previous one's output, so that the selected agents run in order, threading the accumulated state forward:
\begin{equation}
    \text{result} = \alpha_m(\alpha_{m-1}(\ldots\alpha_1(q, \mathcal{S}_0)\ldots, \mathcal{S}_{m-2}), \mathcal{S}_{m-1}) ,
\end{equation}
where $\mathcal{S}_i$ is the state accumulated after agent $\alpha_i$ and every agent sees both the original query and the upstream results. By the coordination analysis of Section~\ref{sec:principles} this is the most expensive pattern in latency, costing the full serial sum $\sum_i \mathbb{E}[\tau_i] + (m-1)\tau_c$, and it is therefore reserved for genuine dependencies. A customer order requiring a credit check is the canonical case: the Finance Agent must assess credit before the Sales Agent commits the order, and the Inventory Agent can reserve stock only once the order is confirmed, so the three steps can be neither reordered nor parallelised.

\emph{Parallel coordination} applies when the selected subtasks are mutually independent, so that they execute concurrently and their results are merged:
\begin{equation}
    \text{result} = \text{aggregate}\big(\alpha_1(q_1, \mathcal{S}) \,\|\, \alpha_2(q_2, \mathcal{S}) \,\|\, \ldots \,\|\, \alpha_m(q_m, \mathcal{S})\big) ,
\end{equation}
where $\|$ denotes concurrent execution. The coordination analysis applies directly: latency falls from the serial sum to $\max_i \mathbb{E}[\tau_i] + \tau_{\text{agg}}$, set by the slowest agent rather than the total, which is the favourable regime whenever the aggregation overhead stays within bound~\eqref{eq:coord}. A whole-business review is typical, in which the Sales, Inventory, and Finance Agents independently report recent orders, stock status, and cash position and a final step synthesises the three into a single answer; because none reads another's output, sequential execution would waste most of the latency budget.

\emph{Hierarchical coordination} addresses queries that do not arrive with their decomposition apparent, where deciding how to divide the work is itself part of the task. The decision is delegated to the ERP Coordinator, which splits the query into sub-queries, routes each to a specialist, and synthesises the returns:
\begin{equation}
    \alpha_{\text{coord}}(q) \rightarrow \{(q_i, \alpha_i)\}_{i=1}^{m} ; \qquad \text{result} = \alpha_{\text{coord}}(\{r_i\}_{i=1}^{m}) .
\end{equation}
This pattern is the most general and subsumes the other two, but it spends an additional planning round on the Coordinator, and is therefore reserved for open-ended requests such as a strategic review whose relevant sub-questions are not known in advance.

\begin{table}[H]
\centering
\caption{The three coordination patterns: applicability, a representative ERP query, and the latency trade-off. Here $\tau_c$ is the per-hand-off overhead and $\tau_{\text{agg}}$ the one-time merge cost of Section~\ref{sec:principles}, and $m$ is the number of participating agents.}
\label{tab:coordination}
\small
\begin{tabular}{lp{3.1cm}p{3.6cm}p{4.3cm}}
\toprule
\textbf{Pattern} & \textbf{Applies when} & \textbf{Representative ERP query} & \textbf{Latency and trade-off} \\
\midrule
Sequential & the subtasks form a dependency chain, each consuming the previous output & order with a credit check: Finance assesses credit, then Sales commits the order, then Inventory reserves stock & highest, at the full serial sum $\sum_i \mathbb{E}[\tau_i] + (m-1)\tau_c$; respects genuine dependencies but permits no overlap \\
\midrule
Parallel & the subtasks are mutually independent & whole-business review: Sales, Inventory, and Finance report concurrently and a merge step synthesises the three & lowest, at $\max_i \mathbb{E}[\tau_i] + \tau_{\text{agg}}$, set by the slowest agent; requires a merge step and only applies without cross-dependencies \\
\midrule
Hierarchical & the decomposition is not known in advance, so splitting the query is itself part of the task & open-ended strategic review whose relevant sub-questions emerge only on inspection & most general and subsumes the other two, at the cost of one extra Coordinator planning round \\
\bottomrule
\end{tabular}
\end{table}

A further provision governs disagreement between specialists, who pursue different objectives and may therefore propose incompatible actions, as when a Sales Agent is willing to accept an order that the Finance Agent flags as exceeding the customer's credit limit. Such conflicts are resolved in three stages. Hard constraints are evaluated before anything else: an action that would violate an invariant such as available stock or a credit limit is rejected outright, regardless of which agent proposed it, because the ERP backend would refuse it in any case. Among the actions that remain, a configurable priority order, by default Finance $>$ Operations $>$ Sales, breaks ties in line with the relative cost of the risks each role guards against. A conflict that still cannot be settled automatically is escalated to a human together with the full context from every participating agent, rather than resolved by an arbitrary default. These stages protect, in turn, correctness, business policy, and human authority over genuine judgement calls.

\subsection{Orchestration Algorithms}
\label{sec:orchestration_algo}

The orchestration function $\Omega$ of Section~\ref{sec:methodology} is realised by three algorithms, presented below alongside the design choices that motivate them. Two artefacts are externalised first, so that the orchestrator's behaviour remains inspectable rather than buried in prompts. The first is the grading rubric against which the reflector scores every candidate output.

This rubric is formalised as a set $\mathcal{G} = \{(d_i, w_i, \theta_i)\}$ of triples, in which $d_i$ is a quality dimension, $w_i \in [0,1]$ its weight with $\sum_i w_i = 1$, and $\theta_i$ a minimum per-dimension threshold on a common $[0,10]$ scale. An output $y$ with per-dimension scores $s_i(y)\in[0,10]$ receives the weighted score $S(y) = \sum_i w_i s_i(y)$ and satisfies the rubric when $s_i(y) \geq \theta_i$ for every $i$ and $S(y) \geq \bar\theta$, where $\bar\theta = \sum_i w_i \theta_i$. For reporting, $S$ is normalised to $[0,1]$. The instantiation used throughout fixes four dimensions, Completeness $(w=0.30,\theta=6)$, Accuracy $(w=0.30,\theta=7)$, Actionability $(w=0.25,\theta=5)$, and Efficiency $(w=0.15,\theta=5)$, giving $\bar\theta = 5.85$, and is applied by the LLM-as-judge described in Section~\ref{sec:grader}; the normalised score $S$ is the quality measure used in the rest of the evaluation.

The second artefact is a \emph{sprint contract} $\mathcal{C} = (\mathbf{a}, \mathbf{d}, t_0)$ that the planner and evaluator agree on before execution, comprising testable acceptance criteria $\mathbf{a} = \{a_1, \ldots, a_m\}$, the expected deliverables $\mathbf{d}$, and a creation timestamp $t_0$. Fixing this agreement up front gives completion a concrete, checkable definition and guards against both premature termination and scope creep. Because the rubric and the contract are data rather than code, they can be revised, audited, and version-controlled independently of the agents that consume them. Swapping a scenario's acceptance criteria, not its implementation, is what lets the same orchestrator serve a new task.

With these artefacts in place, Algorithm~\ref{alg:orchestration} presents the orchestration pipeline that implements the Planner--Executor--Reflector--Responder pattern.

\begin{algorithm}[H]
\caption{LangGraph Orchestration Pipeline}
\label{alg:orchestration}
\begin{algorithmic}[1]
\REQUIRE Query $q$, Enterprise state $\mathcal{S}$, Max iterations $K$
\ENSURE Response $y$ or escalation request
\STATE $\mathbf{plan} \leftarrow \text{PLAN}(q, \mathcal{S})$ \COMMENT{Decompose into subtasks}
\STATE $k \leftarrow 0$
\WHILE{$k < K$}
    \STATE $\mathbf{results} \leftarrow \emptyset$
    \FOR{each subtask $(t_i, \alpha_i) \in \mathbf{plan}$}
        \STATE $r_i \leftarrow \text{EXECUTE}(t_i, \alpha_i, \mathcal{S})$
        \STATE $\mathbf{results} \leftarrow \mathbf{results} \cup \{r_i\}$
    \ENDFOR
    \STATE $(\text{success}, \text{feedback}) \leftarrow \text{REFLECT}(\mathbf{results}, q)$
    \IF{success}
        \RETURN $\text{RESPOND}(\mathbf{results}, q)$
    \ENDIF
    \STATE $\mathbf{plan} \leftarrow \text{REPLAN}(q, \mathcal{S}, \text{feedback})$
    \STATE $k \leftarrow k + 1$
\ENDWHILE
\RETURN $\text{ESCALATE}(q, \mathbf{results})$
\end{algorithmic}
\end{algorithm}

The pipeline is intentionally a bounded loop rather than open-ended agentic recursion. \textsc{Plan} (line~1) converts the query into a typed-tool plan, an ordered set of $(t_i, \alpha_i)$ subtask--agent pairs, so that execution becomes a scheduling problem rather than free-form generation. This is also where the role-scoping of Section~\ref{sec:principles} takes effect, since each subtask is bound to the single agent that owns the relevant tools. \textsc{Execute} (lines~5--8) dispatches each subtask to its owning agent and collects structured results. The decisive design choice is that \textsc{Reflect} (line~9) returns a \emph{pair} $(\text{success}, \text{feedback})$ rather than a single bit: the boolean gates termination, while the feedback carries the information that the subsequent \textsc{Replan} conditions on. That feedback is exactly the decorrelating signal that makes the reflection bound~\eqref{eq:reflbound} non-vacuous, because a replan informed by why the previous attempt failed is less likely to repeat the failure than an independent retry. The loop is capped at $K$ iterations, and the reported configuration permits a single corrective replan: the marginal benefit of further iterations falls off as $(1-p)^k$ while latency and token cost grow linearly, so additional attempts rarely justify their expense. When the cap is reached without an accepted output the query is escalated to a human (line~16) rather than answered with a low-confidence result, which is the orchestration-level expression of the human-in-the-loop principle.

As a concrete illustration, consider a customer that places an order whose value would push its outstanding balance past its credit limit. \textsc{Plan} decomposes the request and binds each subtask to its owner: a credit assessment for the Finance Agent, an availability check for the Inventory Agent, and order creation for the Sales Agent. \textsc{Execute} dispatches these, and the Inventory Agent confirms stock while the Finance Agent returns a structured result reporting that the order would exceed the credit limit by a specific amount. \textsc{Reflect} finds no tool-execution error, but the acceptance criterion that the order be committable is unmet, so it returns $(\text{False}, \text{feedback})$ with the credit shortfall as the feedback rather than a bare rejection. Conditioned on that feedback, \textsc{Replan} does not retry the same commit; it revises the plan to the options the shortfall admits, such as a partial order within the available limit or a commit contingent on prepayment, and because overriding a credit limit is a high-risk action the revised plan routes the decision to a human reviewer through the safety harness rather than executing it autonomously. \textsc{Respond} then returns the recommended options together with the credit gap that motivated them. The episode makes the division of labour concrete: the boolean halts a commit that would violate a business invariant, while the feedback is what lets the next plan address the specific reason for the halt instead of repeating it.

Assigning agents to subtasks, the function performed within \textsc{Plan}, rests on a two-stage intent-routing procedure formalised in Algorithm~\ref{alg:routing}.

\begin{algorithm}[H]
\caption{Two-Stage Intent Routing}
\label{alg:routing}
\begin{algorithmic}[1]
\REQUIRE Query $q$, Agent set $\mathcal{A} = \{\alpha_1, \ldots, \alpha_n\}$, Threshold $\theta$
\ENSURE Active agent set $\mathcal{A}^* \subseteq \mathcal{A}$
\STATE \COMMENT{Stage 1: Primary classification}
\STATE $\mathbf{d} \leftarrow \text{LLM}(q, \{\text{desc}(\alpha_i)\}_{i=1}^n)$ \COMMENT{Role descriptions}
\STATE $\alpha^* \leftarrow \arg\max_{\alpha_i} \mathbf{d}_i$
\STATE $\mathcal{A}^* \leftarrow \{\alpha^*\}$
\STATE \COMMENT{Stage 2: Multi-agent analysis}
\STATE $\mathbf{c} \leftarrow \text{COMPLEXITY}(q)$ \COMMENT{Query complexity score}
\IF{$\mathbf{c} > c_{\text{threshold}}$}
    \FOR{each $\alpha_i \in \mathcal{A} \setminus \{\alpha^*\}$}
        \IF{$\mathbf{d}_i > \theta$}
            \STATE $\mathcal{A}^* \leftarrow \mathcal{A}^* \cup \{\alpha_i\}$
        \ENDIF
    \ENDFOR
\ENDIF
\RETURN $\mathcal{A}^*$
\end{algorithmic}
\end{algorithm}

Routing is split into two stages because the two failure modes it must avoid pull in opposite directions. Stage~1 (lines~2--4) always commits to exactly one primary agent, the $\arg\max$ of an LLM classification over the agents' role descriptions. It guarantees that even an ambiguous query is served by some specialist rather than stalling. Stage~2 (lines~5--9) admits further agents only when a separate complexity estimate exceeds a threshold, and then only those scoring above $\theta$. The asymmetry is deliberate: invoking a second agent is cheap insurance on a genuinely cross-functional query but pure overhead on a simple one, so the default is a single agent and breadth is added only on evidence of cross-functional structure. Both thresholds, $c_{\text{threshold}}$ and $\theta$, are exposed as configuration rather than hard-coded, which is what allows the precision--recall trade-off to be tuned per deployment without touching the agents themselves.

The \textsc{Reflect} step invoked at line~9 of the pipeline carries the self-correction logic, detailed in Algorithm~\ref{alg:reflection}.

\begin{algorithm}[H]
\caption{Self-Correction via Reflection}
\label{alg:reflection}
\begin{algorithmic}[1]
\REQUIRE Execution results $\mathbf{R}$, Original query $q$, Grading rubric $\mathcal{G}$, Acceptance criteria $\mathbf{a}$
\ENSURE (success: bool, feedback)
\STATE \COMMENT{Check tool execution success}
\FOR{each $r_i \in \mathbf{R}$}
    \IF{$r_i.\text{status} = \text{ERROR}$}
        \RETURN $(\text{False}, r_i)$ \COMMENT{return the failed result as feedback}
    \ENDIF
\ENDFOR
\STATE \COMMENT{Check response completeness}
\STATE $\text{coverage} \leftarrow \text{COVERAGE}(q, \mathbf{R})$ \COMMENT{LLM-as-judge coverage check}
\IF{coverage $< 0.8$}
    \RETURN $(\text{False}, \text{coverage})$ \COMMENT{coverage gaps as feedback}
\ENDIF
\STATE \COMMENT{Check constraint satisfaction}
\FOR{each acceptance criterion $c \in \mathbf{a}$}
    \IF{$\neg\text{SATISFIES}(\mathbf{R}, c)$}
        \RETURN $(\text{False}, c)$ \COMMENT{the unmet criterion as feedback}
    \ENDIF
\ENDFOR
\RETURN $(\text{True}, \emptyset)$
\end{algorithmic}
\end{algorithm}

The reflector applies three checks in increasing order of cost and short-circuits on the first failure. The cheapest is a deterministic scan for tool exceptions (lines~2--6): a failed Odoo call is an unambiguous, machine-detectable error, so there is no reason to spend an LLM call to notice it, and the offending tool and message are returned directly as feedback. Only when execution was clean does the reflector invoke the LLM-as-judge, first for a coverage check against the query (lines~8--11) and then for a pass/fail test of each acceptance criterion in the contract (lines~12--16), with the grading rubric $\mathcal{G}$ supplying the dimensions and thresholds. Ordering the deterministic check first keeps most mechanical failures away from the model, and emitting a specific feedback string rather than a bare reject is what gives \textsc{Replan} something to act on. Separating this evaluator from the generator is the second design principle made operational: an agent that grades its own work exhibits a systematic positive bias, so a distinct judge applying an external rubric supplies the signal that self-assessment would not.

\subsection{Tool Execution, Safety Harness, and Skills}

Every tool that bridges an agent to the ERP follows a standardised execution pattern that enforces consistency, safety, and auditability. This pattern is also where the safety harness intercepts each write, the mechanism that ensures no unreviewed action reaches the backend.

Tools are partitioned by operational character into three categories. \emph{Read} tools query data without modifying it (searching orders, checking inventory levels, retrieving customer records) and execute without approval. \emph{Write} tools create or modify data (creating sales orders, updating customer records, adjusting inventory) and may require approval depending on magnitude or category. \emph{Action} tools trigger business-process transitions (confirming orders, posting invoices, approving requests) and typically represent points of no return that mandate the strictest level of oversight.

This harness, which wraps every tool call and constitutes the architecture's principal safety mechanism, decides whether a write needs approval and from whom by assigning each candidate action one of four risk levels. \textsc{Low}-risk actions, such as reads and small routine writes, execute automatically. \textsc{Medium}-risk actions trigger agent-to-agent review, in which a second specialist validates the action before it commits according to a fixed cross-validation matrix. A sales order, for instance, is checked by Finance for credit and exposure and by Inventory for availability before it is confirmed. \textsc{High}- and \textsc{Critical}-risk actions, such as large payments or irreversible cancellations, require human approval and cannot be auto-executed under any agent reasoning.

The risk level is computed from configurable triggers rather than fixed per tool: a monetary threshold (orders above \$10{,}000 in the reference configuration), an operation category that is always sensitive (e.g., deleting a customer), or an anomaly flag raised when an action departs from historical patterns. Authorised users may pre-approve specific operations so that routine work is not needlessly interrupted, and a feedback loop adjusts the thresholds over time from observed outcomes, tightening them where auto-executed actions later proved costly and relaxing them where human review proved unnecessary. Every gated decision, recorded with its originating agent, inputs, risk level, and approver where applicable, is written to the audit trail, so that the harness is not only a gate but the record that makes each autonomous action accountable after the fact. Gating writes in this way is what guarantees that an unreviewed action cannot reach the ERP, the property on which the rest of the paper relies when it treats the system as safe to run autonomously.

The tools the harness governs are partitioned by operational character into three categories. \emph{Read} tools query data without modifying it (searching orders, checking inventory levels, retrieving customer records) and execute without approval. \emph{Write} tools create or modify data (creating sales orders, updating customer records, adjusting inventory) and may require approval depending on magnitude or category. \emph{Action} tools trigger business-process transitions (confirming orders, posting invoices, approving requests) and typically represent points of no return that mandate the strictest level of oversight.

Each tool implementation follows a standardised pattern, given in Algorithm~\ref{alg:tool_pattern}:

\begin{algorithm}[H]
\caption{Standardised Tool Execution Pattern}
\label{alg:tool_pattern}
\begin{algorithmic}[1]
\REQUIRE Tool parameters, user context, and session state
\ENSURE Execution result or error
\STATE \textbf{Validate} parameters against schema
\IF{validation fails}
    \RETURN error with specific feedback
\ENDIF
\STATE \textbf{Check} business rules and constraints
\IF{constraints violated}
    \RETURN error explaining constraint
\ENDIF
\STATE \textbf{Check} approval requirements
\IF{approval required AND not pre-approved}
    \STATE Queue for approval
    \RETURN pending approval status
\ENDIF
\STATE \textbf{Execute} ERP API call within transaction
\IF{execution fails}
    \STATE Rollback any partial changes
    \RETURN error with details
\ENDIF
\STATE \textbf{Log} execution for audit trail
\RETURN success result
\end{algorithmic}
\end{algorithm}

The order of these steps is itself the safeguard. Parameter validation and business-rule checks run first, so malformed or non-compliant calls are rejected before any reviewer or transaction is involved. The approval gate sits between checking and executing: a tool that requires approval and has not been pre-approved returns a pending status and is never executed, which separates the decision to act from the act itself. Execution then runs inside a transaction, so a failure rolls back cleanly and leaves no partial write, and every successful call is appended to an audit log. Because all $46$ tools share this single pattern rather than reimplementing it, the safety properties hold uniformly: whatever an agent infers, a write that fails validation, breaks a business rule, or lacks a required approval cannot reach Odoo.

Beyond individual tools, operational know-how that belongs to no single tool, such as the steps of a month-end close or a supplier-onboarding workflow, is captured as \emph{skills} rather than embedded in agent code. Each skill is a self-contained document stating when it applies and how to carry it out. To keep prompts small, skills are disclosed progressively: at startup only their names and one-line descriptions are loaded into the system prompt, and the full body of a skill is injected into context only when an incoming query matches its description. The standing context cost is therefore independent of the size of the skill library, so adding a workflow does not enlarge every prompt. This mirrors role-scoping one level up, with an agent holding only the knowledge a task actually requires, and it keeps workflows externalised and inspectable in the same spirit as the grading criteria and contracts of Section~\ref{sec:orchestration_algo}: a new procedure is added as a skill document rather than as code.

\section{Experimental Design}
\label{sec:experiments}

\subsection{Design Overview}
\label{sec:grader}

The evaluation has five components, summarised in Table~\ref{tab:experiments}. Experiment~E1 is a scenario-based task suite of $36$ items across six functional categories that characterises per-domain behaviour and routing. Experiment~E2 is a head-to-head comparison of six orchestration paradigms on a $15$-scenario enterprise-crisis benchmark (five cash-flow, five customer-dispute, five supplier-disruption). Experiment~E3 is a 365-day agent-in-the-loop simulation driven by stochastic demand and supply, compared against rule-based RPA and no-intervention baselines. An ablation study (E4) and a harness-enhancement study (E5) isolate the contribution of individual components. All generation cells use GPT-4o-mini at temperature $0.1$. The experimental infrastructure is provider-agnostic to support replication on alternative LLM backends.

\begin{table}[H]
\centering
\caption{Experiments and the claim each is designed to test.}
\label{tab:experiments}
\small
\begin{tabular}{lp{9.5cm}}
\toprule
\textbf{Experiment} & \textbf{Claim tested} \\
\midrule
E1 (scenarios, tea) & Role-aligned decomposition handles diverse enterprise tasks; routing accuracy and task completion are measurable. \\
E2 (orchestration, $15$ crisis scenarios) & Graph-based Planner--Executor--Reflector--Responder orchestration is competitive with the best multi-agent baselines on cross-functional crisis scenarios and significantly outperforms single-step reasoning. \\
E3 (year-long sim) & Sustained autonomous operation is feasible; LLM-agent decisions improve operational quality over rule-based RPA under identical demand. \\
E4 (ablation) & Predictions P1--P3 hold: role decomposition, reflection, and planner each contribute measurably. \\
E5 (harness V1\,$\to$\,V2) & Externalised grading criteria and sprint contracts improve quality on the most complex scenarios. \\
\bottomrule
\end{tabular}
\end{table}

For E1 and the $15$-scenario crisis suite, the evaluation adopts an LLM-as-judge protocol. Each scenario ships with a machine-readable oracle comprising (a) a list of acceptance criteria and (b) an expected tool set. After the system produces a response, a separately-configured LLM (distinct from the generator, running at temperature $0.0$) receives the prompt, the acceptance-criterion list, and the response, and returns per-dimension scores on the $1$--$10$ rubric (Completeness, Accuracy, Actionability, Efficiency) together with a pass/fail bit per acceptance criterion. The weighted aggregate uses the weights specified in Section~\ref{sec:orchestration_algo}. The judge LLM is held constant across cells of a given experiment so that rating variability comes from the generator, not the judge. This protocol is a scalable proxy for expert assessment rather than a gold standard: separating the judge from the generator, so that no model grades its own output, controls the self-evaluation bias to which single-agent grading is prone, but the resulting scores remain an automated approximation of response quality rather than a calibrated substitute for human expert judgement, a limitation revisited in Section~\ref{sec:discussion}.

\subsection{Experimental Setting}

The tea-trading company used throughout is a controlled small-to-medium enterprise (SME) benchmark rather than a representative sample of ERP deployments: it is deliberately compact and fully observable, so that system behaviour is reproducible and attributable, and it is not meant to stand in for the scale or regulatory complexity of larger installations. It does, however, exercise the four functional modules that dominate day-to-day operation in most trading firms, namely Sales, Inventory, Purchasing, and Accounting. Concretely, the simulation represents \emph{Jade Garden Tea House}, a specialty tea trading company with $26$ products across four categories (bulk tea, premium packaged tea, brewing accessories, gift sets), $7$ customer accounts spanning enterprise hotels, wholesale distributors, and retail cafes, $5$ suppliers with heterogeneous lead times, reliability, and specialisations, and an initial cash position of \$1{,}000{,}000. Goods flow in days rather than weeks and payment is typically on delivery, which stresses the service-level dimension of the system but does not exercise complex working-capital dynamics.

\begin{figure}[H]
    \centering
    \includegraphics[width=0.8\textwidth]{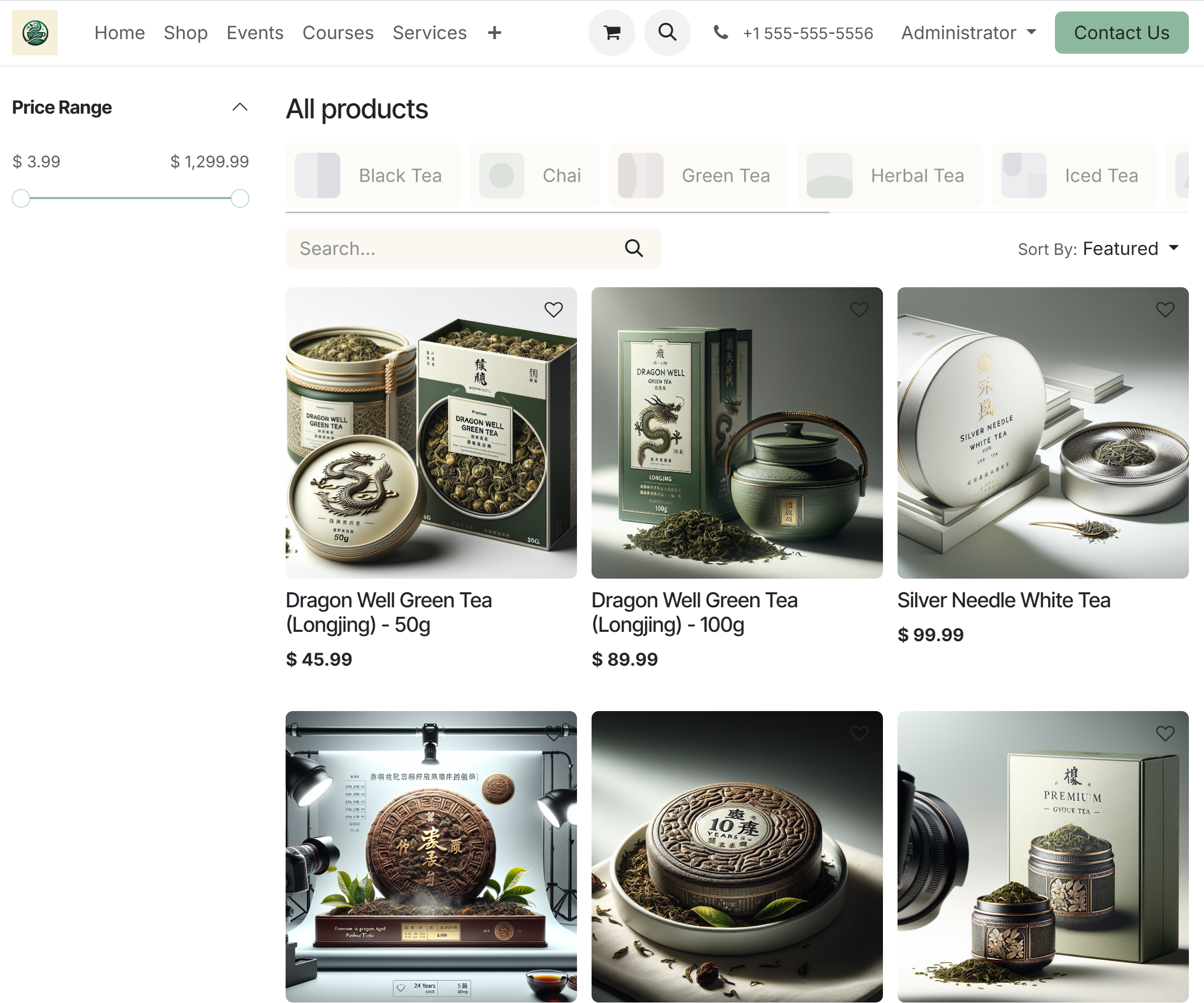}
    \caption{Online shop of the Jade Garden Tea House tea trading company.}
    \label{fig:shop}
\end{figure}

Figure \ref{fig:shop} shows the webpage of the online shop of the the Jade Garden Tea House tea trading company.
The experimental system runs on Ubuntu 22.04 LTS, Odoo 19 Community Edition with PostgreSQL, and Python 3.10. The reported results use OpenAI GPT-4o-mini at temperature $0.1$ for generation while the LLM-as-judge grader (Section~\ref{sec:grader}) runs at temperature $0.0$. 

\subsection{Evaluation Scenarios}

E1 uses a scenario suite organised into six functional categories (supplier selection, demand forecasting, pricing decisions, anomaly response, customer service, and multi-agent collaboration), with six scenario specifications per category, yielding a $36$-item design. Each specification is a triple $(\mathrm{prompt},\mathrm{initial\ state},\mathrm{oracle\ acceptance\ criteria})$, with identifiers of the form $\{S,D,P,A,C,M\}\,k$ for $k\in\{1,\ldots,6\}$. The categories are designed to exercise distinct competencies of the architecture: \emph{Supplier selection} probes the Purchasing Agent's vendor evaluation; \emph{Demand forecasting} probes the Inventory Agent's reorder reasoning; \emph{Pricing decisions} probe the Sales Agent's discount and margin reasoning; \emph{Anomaly response} probes investigation and corrective action across the Coordinator and the relevant specialist; \emph{Customer service} probes inquiry handling against order, product, and history data; and \emph{Multi-agent collaboration} probes cross-functional workflows that necessarily span at least two agents. Within each category, the six specifications are arranged in approximately ascending difficulty so that the suite covers the operating envelope of each competency. Table~\ref{tab:e1_scenarios} summarises every specification by category, identifier, brief description, primary agents required, and complexity tier; complexity is graded \emph{S} (single-domain, mechanical), \emph{M} (multi-step within one domain or two-domain), or \emph{H} (multi-domain with judgement). The E1 run used for the main results table executes each specification once per orchestration configuration with a fixed seed; the multi-run statistical analysis in Section~\ref{sec:stats} is a post-hoc stability study over a $10$-scenario stratified subsample, marked $\bullet$ in the rightmost column.

\begingroup
\small
\setlength{\LTleft}{\fill}
\setlength{\LTright}{\fill}
\begin{longtable}{llp{6.0cm}lcc}
\caption{E1 scenario suite ($N=36$). \emph{Agents}: ER = ERP Coordinator, S = Sales, I = Inventory, P = Purchasing, F = Finance. \emph{Complexity}: S (single-domain), M (medium), H (hard, multi-domain with judgement). The rightmost column marks the $10$ scenarios in the stability subsample used in Section~\ref{sec:stats}.}
\label{tab:e1_scenarios}\\
\toprule
\textbf{Category} & \textbf{ID} & \textbf{Description} & \textbf{Agents} & \textbf{Cmplx.} & \textbf{Stab.} \\
\midrule
\endfirsthead
\multicolumn{6}{l}{\footnotesize\emph{Table~\ref{tab:e1_scenarios} (continued)}} \\
\toprule
\textbf{Category} & \textbf{ID} & \textbf{Description} & \textbf{Agents} & \textbf{Cmplx.} & \textbf{Stab.} \\
\midrule
\endhead
\midrule
\multicolumn{6}{r}{\footnotesize\emph{continued on next page}} \\
\endfoot
\bottomrule
\endlastfoot
\multirow{6}{*}{\shortstack[l]{Supplier\\selection}}
 & S1 & Basic supplier comparison for a single product & P & S & $\bullet$ \\*
 & S2 & Multi-criteria evaluation (price, lead time, reliability) & P & M & \\*
 & S3 & Urgent procurement with availability constraints & P, I & M & $\bullet$ \\*
 & S4 & New supplier evaluation with limited history & P & M & \\*
 & S5 & Supplier consolidation analysis & P, F & H & $\bullet$ \\*
 & S6 & Risk-adjusted sourcing decision & P, F & H & \\
\midrule
\multirow{6}{*}{\shortstack[l]{Demand\\forecasting}}
 & D1 & Basic reorder-point calculation & I & S & $\bullet$ \\*
 & D2 & Seasonal demand adjustment & I, S & M & \\*
 & D3 & New-product-launch inventory planning & I, S & M & $\bullet$ \\*
 & D4 & Promotional demand anticipation & I, S & M & \\*
 & D5 & Multi-location inventory optimisation & I & H & \\*
 & D6 & Demand--supply mismatch resolution & I, P & H & \\
\midrule
\multirow{6}{*}{\shortstack[l]{Pricing\\decisions}}
 & P1 & Standard discount approval & S & S & \\*
 & P2 & Volume-based pricing negotiation & S, F & M & $\bullet$ \\*
 & P3 & Competitive response pricing & S & M & \\*
 & P4 & Customer-specific pricing evaluation & S, F & M & $\bullet$ \\*
 & P5 & Promotional-pricing impact analysis & S, F & H & \\*
 & P6 & Margin-protection decision & S, F & H & \\
\midrule
\multirow{6}{*}{\shortstack[l]{Anomaly\\response}}
 & A1 & Inventory-discrepancy investigation & I, ER & M & $\bullet$ \\*
 & A2 & Unusual transaction-pattern response & F, ER & M & \\*
 & A3 & Quality-issue escalation & I, S & M & \\*
 & A4 & Delivery-failure recovery & P, S & H & \\*
 & A5 & Payment-anomaly resolution & F & H & \\*
 & A6 & Multi-system inconsistency diagnosis & ER, all & H & \\
\midrule
\multirow{6}{*}{\shortstack[l]{Customer\\service}}
 & C1 & Order-status inquiry & S & S & \\*
 & C2 & Product-availability check with alternatives & S, I & M & $\bullet$ \\*
 & C3 & Return / exchange processing & S, I, F & M & \\*
 & C4 & Complaint handling and resolution & S, ER & M & \\*
 & C5 & Account inquiry with history synthesis & S, F & M & \\*
 & C6 & Complex multi-item order processing & S, I, F & H & \\
\midrule
\multirow{6}{*}{\shortstack[l]{Multi-agent\\collaboration}}
 & M1 & Order-to-cash workflow & S\,$\to$\,I\,$\to$\,F & H & $\bullet$ \\*
 & M2 & Procure-to-pay workflow & P\,$\to$\,I\,$\to$\,F & H & \\*
 & M3 & Issue resolution requiring multiple domains & ER, S, I, F & H & \\*
 & M4 & Business review synthesising all functions & ER, all & H & \\*
 & M5 & Exception handling spanning departments & ER, all & H & \\*
 & M6 & Strategic decision with enterprise-wide implications & ER, all & H & \\
\end{longtable}
\endgroup

\subsection{Evaluation Metrics}

Four metrics quantify system behaviour across the experiments and are summarised in Table~\ref{tab:metrics}. The \emph{task completion rate} (TCR) is a binary measure of whether the system successfully completed the requested task,
\begin{equation}
    \text{TCR} = \frac{\text{Successfully completed tasks}}{\text{Total tasks attempted}} \times 100\%,
\end{equation}
where a task counts as complete when the requested action was executed in the ERP system, the execution achieved the intended business outcome, and no errors or inconsistencies resulted. The \emph{agent routing accuracy} (ARA) is the percentage of queries routed to the appropriate agents,
\begin{equation}
    \text{ARA} = \frac{\text{Correctly routed queries}}{\text{Total queries}} \times 100\%,
\end{equation}
with correctness determined by comparing the system's routing against the agents specified in each scenario's oracle. \emph{Response latency} (RL) is the end-to-end time from query submission to response delivery, measured in milliseconds and inclusive of orchestration overhead in multi-agent scenarios. \emph{Token and cost efficiency} (TCE) records the total tokens consumed per task and the corresponding API cost at current pricing (\$0.00015 per 1K input tokens and \$0.0006 per 1K output tokens for GPT-4o-mini).

\begin{table}[H]
\centering
\caption{Evaluation metrics: abbreviations and definitions.}
\label{tab:metrics}
\small
\begin{tabular}{lcp{7.7cm}}
\toprule
\textbf{Metric} & \textbf{Abbr.} & \textbf{Meaning} \\
\midrule
Task completion rate & TCR & Percentage of attempted tasks executed in the ERP with the intended business outcome and no error or inconsistency. \\
Agent routing accuracy & ARA & Percentage of queries dispatched to the agent(s) specified by the scenario oracle. \\
Response latency & RL & End-to-end time (ms) from query submission to response delivery, inclusive of orchestration overhead. \\
Token and cost efficiency & TCE & Tokens consumed per task and the corresponding API cost at GPT-4o-mini pricing. \\
\bottomrule
\end{tabular}
\end{table}

\section{Results}
\label{sec:results}

\subsection{Task-Level Performance}
\label{sec:stats}

Experiment E1 tests whether the role-aligned architecture handles the full breadth of enterprise tasks. Table~\ref{tab:results_detailed} presents detailed results across all scenario categories.

\begin{table}[H]
\centering
\caption{Detailed results by scenario category on the controlled $36$-scenario E1 suite. TCR is task completion in the transactional sense of Section~\ref{sec:experiments}, not a quality score.}
\label{tab:results_detailed}
\begin{tabular}{lcccc}
\toprule
\textbf{Category} & \textbf{TCR} & \textbf{RL} & \multicolumn{2}{c}{\textbf{TCE}} \\
\cmidrule(lr){4-5}
 & & (ms) & Tokens & Cost (\$) \\
\midrule
Supplier Selection & 100\% & 1,150 & 385 & 0.0012 \\
Demand Forecasting & 100\% & 1,280 & 410 & 0.0013 \\
Pricing Decisions & 100\% & 1,050 & 375 & 0.0011 \\
Anomaly Response & 100\% & 1,450 & 445 & 0.0014 \\
Customer Service & 100\% & 890 & 350 & 0.0010 \\
Multi-Agent & 100\% & 2,320 & 520 & 0.0018 \\
\midrule
\textbf{Overall} & \textbf{100\%} & \textbf{1,190} & \textbf{401} & \textbf{0.0013} \\
\bottomrule
\end{tabular}
\end{table}

Two observations summarise the table. The system completes every scenario in the E1 suite under the operational definition introduced in Section~\ref{sec:experiments}. The requested action is executed in Odoo, the post-condition implied by the prompt is satisfied, and no transactional error is raised. Completion in this sense does not imply that every response is the best possible response, only that it is transactionally correct. Mean per-query latency of $1.19$\,s and a mean of $401$ tokens per query place the system within the range of interactive use. Multi-agent scenarios that require explicit coordination roughly double the latency ($2.32$\,s) owing to sequential agent invocations and orchestration overhead.

The numbers in Table~\ref{tab:results_detailed} come from single-seed runs of the full 36-scenario design. To quantify stochastic variability of the LLM substrate, a stability study was conducted in which the 10-scenario stratified subsample defined in Section~\ref{sec:experiments} was executed five times with different random seeds (temperature $0.1$; seeds $\{s_1,\ldots,s_5\}$). All confidence intervals, $t$-tests, and effect sizes reported below are computed over this stability subsample and are not claimed to generalise to scenarios outside it.

Table~\ref{tab:confidence} reports 95\% confidence intervals from the stability subsample ($n=5$ runs $\times$ $10$ scenarios $= 50$ observations per metric, except TCR which is Clopper--Pearson on pooled binary outcomes).

\begin{table}[H]
\centering
\caption{95\% confidence intervals for key metrics (n=5 runs per scenario)}
\label{tab:confidence}
\begin{tabular}{lcc}
\toprule
\textbf{Metric} & \textbf{Mean} & \textbf{95\% CI} \\
\midrule
TCR & 100\% & [100\%, 100\%] \\
ARA & 98\% & [95.2\%, 100\%] \\
RL (ms) & 1,190 & [1,045, 1,335] \\
TCE (tokens/task) & 401 & [372, 430] \\
\bottomrule
\end{tabular}
\end{table}

Paired two-sided $t$-tests on the stability subsample ($n=10$ scenario pairs, each aggregated across five seeds) compare the multi-agent (MA) orchestration against the single-agent (SA) baseline; Table~\ref{tab:ma_vs_sa} reports the comparison together with Cohen's $d_z$ effect sizes on the same subsample.

\begin{table}[H]
\centering
\caption{Paired comparison of multi-agent (MA) and single-agent (SA) configurations on the stability subsample ($n=10$ scenario pairs, five seeds each). $t(9)$ is the paired two-sided $t$-statistic; $d_z$ is Cohen's effect size for paired observations.}
\label{tab:ma_vs_sa}
\small
\begin{tabular}{lccccc}
\toprule
\textbf{Metric} & \textbf{MA} & \textbf{SA} & $\boldsymbol{t(9)}$ & $\boldsymbol{p}$ & $\boldsymbol{d_z}$ \\
\midrule
TCR & $100\%$ & $83\%$ & $4.82$ & $<\!0.001$ & $1.24$ \\
RL (s) & $1.19$ & $2.45$ & $-6.14$ & $<\!0.001$ & $1.89$ \\
TCE (tokens) & $401$ & $850$ & $-8.92$ & $<\!0.001$ & $2.15$ \\
\bottomrule
\end{tabular}
\end{table}

All three differences survive a Holm--Bonferroni correction at $\alpha = 0.01$, and all three effect sizes exceed the conventional $0.8$ threshold for a large effect \citep{wohlin2012experimentation}. Because the stability subsample is stratified across categories but still small, the $p$-values and effect magnitudes are reported as supporting rather than decisive evidence; the pattern agrees with the structural prediction P1 of the decomposition bound~\eqref{eq:decomp}.

The intent routing mechanism, which underlies these task-completion results, achieved 98\% accuracy across all queries. Table~\ref{tab:routing} reports routing accuracy for each intended agent.

\begin{table}[H]
\centering
\caption{Agent routing accuracy by intended agent. The \emph{Misrouted to} column gives the agent that received the misrouted queries.}
\label{tab:routing}
\begin{tabular}{lcccc}
\toprule
\textbf{Agent} & \textbf{Queries} & \textbf{Correct} & \textbf{Accuracy} & \textbf{Misrouted to} \\
\midrule
ERP Coordinator & 8 & 8 & 100\% & -- \\
Sales & 12 & 12 & 100\% & -- \\
Inventory & 10 & 10 & 100\% & -- \\
Purchasing & 8 & 8 & 100\% & -- \\
Finance & 9 & 9 & 100\% & -- \\
Multi-Agent & 6 & 5 & 83\% & Single agent \\
\midrule
\textbf{Total} & \textbf{53} & \textbf{52} & \textbf{98\%} & \\
\bottomrule
\end{tabular}
\end{table}

Routing errors occurred primarily in multi-agent scenarios that were initially handled by a single agent before orchestration recognised the need for coordination. These errors were recoverable: the system still completed tasks successfully, though with suboptimal routing.

\begin{figure}[H]
    \centering
    \includegraphics[width=\textwidth]{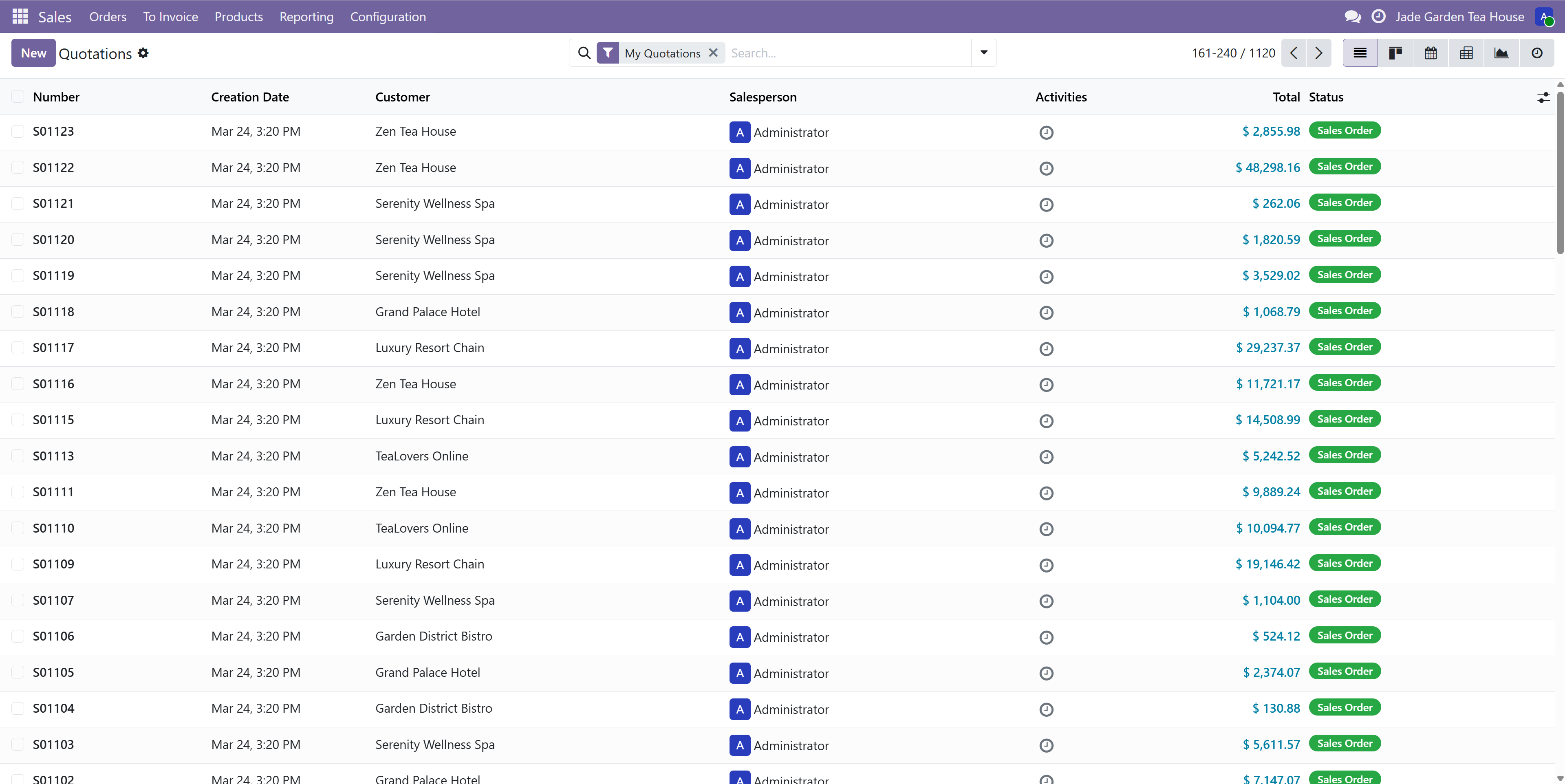}
    \caption{Online order data in Odoo database.}
    \label{fig:order}
\end{figure}

\begin{figure}[H]
    \centering
    \includegraphics[width=\textwidth]{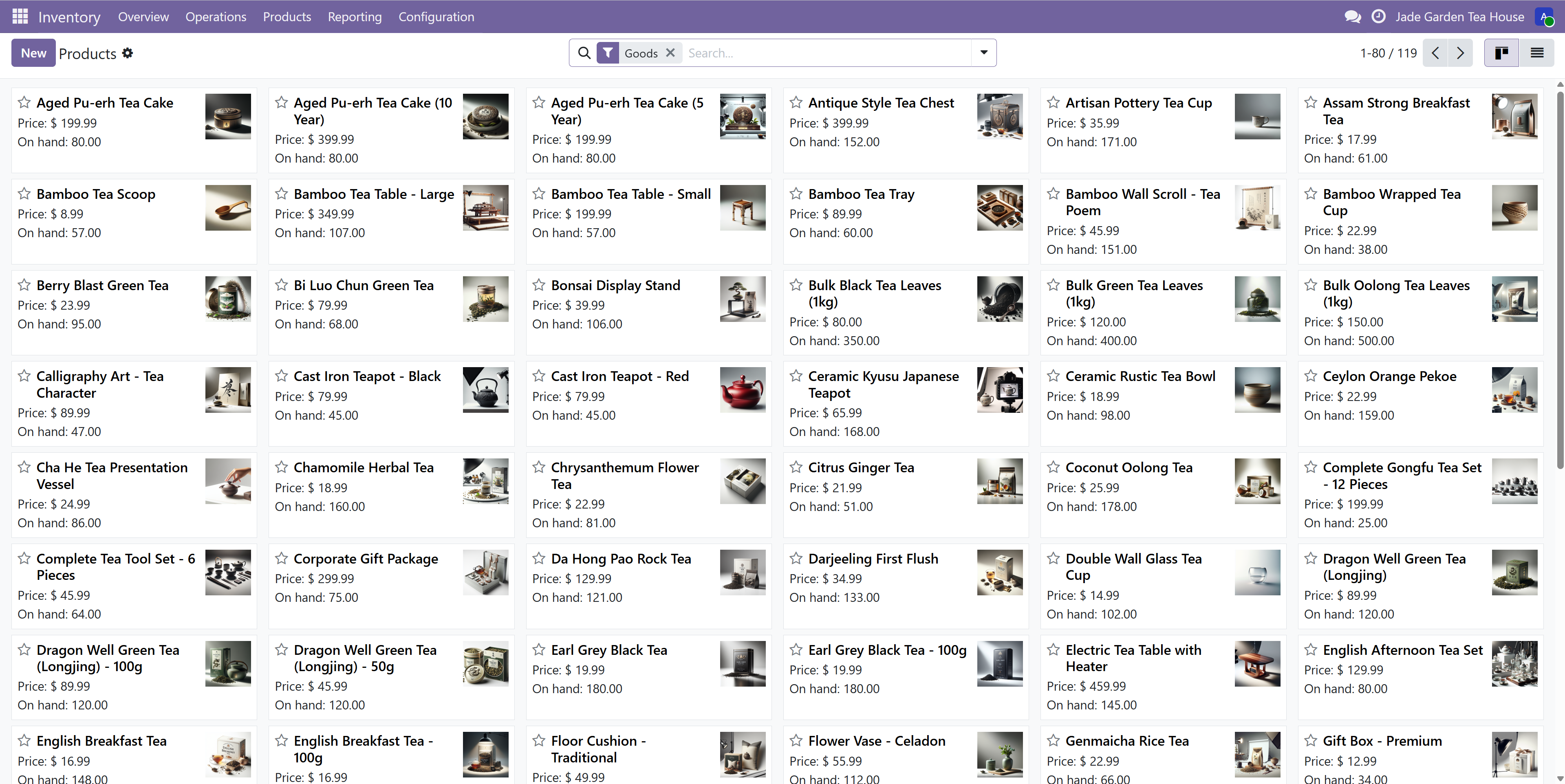}
    \caption{Online inventory data in Odoo database.}
    \label{fig:inventory}
\end{figure}

From the perspective of application, the proposed method worked with Odoo well. For instance, Figure \ref{fig:order} and Figure \ref{fig:inventory} are the data about finished orders and well-maintained inventory, respectively.

\subsection{Long-Term Operation and Baseline Comparison}

To evaluate sustained autonomous operation beyond individual scenarios, and to test whether agentic reasoning improves operational quality over conventional automation, a 365-day business simulation was conducted in which the AI system operated the enterprise with minimal human intervention.

The simulation environment generates realistic business events based on configurable behavioral models:

Customers follow defined ordering patterns parameterised by frequency, typical order size, and product preferences, with stochastic variation about the means; enterprise customers place weekly large-volume orders, while retail customers order monthly at smaller quantities. Suppliers respond to purchase orders with configurable lead times and reliability rates, and a $10\%$ probability of delivery delay is injected to exercise adaptive response. Payment occurs at the contractual terms of each customer (15--60 days) with $95\%$ on-time probability. Occasional disruptions (customer complaints, quality issues, unusual demand spikes) are sampled from low-probability events to exercise exception handling.

Table~\ref{tab:yearly_results} summarizes the 365-day simulation outcomes.

\begin{table}[H]
\centering
\caption{365-day autonomous operation results}
\label{tab:yearly_results}
\begin{tabular}{lrp{6cm}}
\toprule
\textbf{Metric} & \textbf{Value} & \textbf{Interpretation} \\
\midrule
\multicolumn{3}{l}{\textit{Financial Performance}} \\
Starting Capital & \$944,220 & After initial inventory investment \\
Ending Cash & \$3,184,905 & 237\% growth from starting position \\
Revenue Collected & \$3,365,944 & Total customer payments received \\
Net Profit & \$2,240,685 & Cash increase over simulation period \\
Accounts Receivable & \$206,513 & Outstanding customer balances (healthy) \\
\midrule
\multicolumn{3}{l}{\textit{Operational Metrics}} \\
Orders Received & 639 & Average 1.75 orders per day \\
Orders Fulfilled & 266 & Payments collected (others in AR) \\
AI Decisions & 1,203 & Total autonomous actions taken \\
\midrule
\multicolumn{3}{l}{\textit{Service Quality}} \\
Stockout Events & 0 & Perfect product availability \\
Customer Complaints & 4 & 0.6\% complaint rate \\
Supplier Delays & 30 & All handled without customer impact \\
\midrule
\multicolumn{3}{l}{\textit{Efficiency}} \\
Simulation Runtime & 78.15 min & Real computation time for 365 days \\
Decisions per Day & 3.3 & Average AI actions per simulated day \\
\bottomrule
\end{tabular}
\end{table}

\begin{figure}[H]
    \centering
    \includegraphics[width=\textwidth]{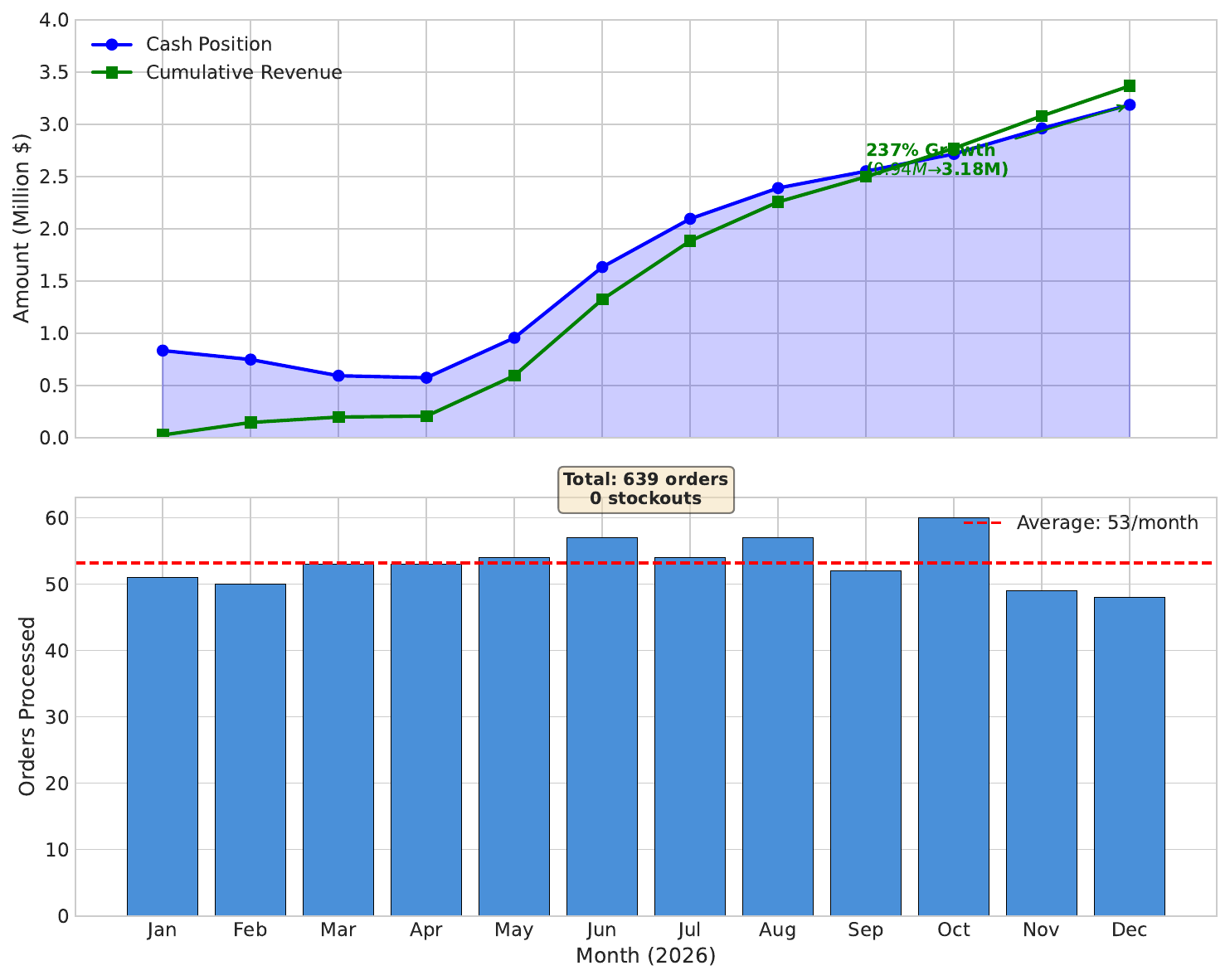}
    \caption{365-day autonomous simulation results. Top: Cash position and cumulative revenue trends showing 237\% capital growth. Bottom: Monthly order volume with 639 total orders processed. The system maintained zero stockouts throughout the simulation period.}
    \label{fig:yearly_simulation}
\end{figure}

Figure~\ref{fig:yearly_simulation} visualizes the 365-day simulation results, showing cash position growth, cumulative revenue, and monthly order volume.

Table~\ref{fig:monthly_detail} shows monthly performance demonstrating consistent growth throughout the year.

\begin{table}[H]
\centering
\caption{Monthly performance progression during 365-day simulation}
\label{fig:monthly_detail}
\begin{tabular}{lrrrr}
\toprule
\textbf{Month} & \textbf{Orders} & \textbf{Revenue} & \textbf{Cash} & \textbf{Growth} \\
\midrule
January & 51 & \$25,850 & \$834,514 & -- \\
February & 50 & \$145,829 & \$748,210 & -10.3\% \\
March & 53 & \$198,499 & \$592,623 & -20.8\% \\
April & 53 & \$207,169 & \$574,018 & -3.1\% \\
May & 54 & \$593,859 & \$955,621 & +66.5\% \\
June & 57 & \$1,322,665 & \$1,633,096 & +70.9\% \\
July & 54 & \$1,882,700 & \$2,094,561 & +28.3\% \\
August & 57 & \$2,255,695 & \$2,388,981 & +14.1\% \\
September & 52 & \$2,496,767 & \$2,550,150 & +6.7\% \\
October & 60 & \$2,769,324 & \$2,714,847 & +6.5\% \\
November & 49 & \$3,078,701 & \$2,959,670 & +9.0\% \\
December & 48 & \$3,365,944 & \$3,184,905 & +7.6\% \\
\bottomrule
\end{tabular}
\end{table}

The early months show cash decline as the system invested in inventory to meet growing demand. From May onward, revenue collection overtook expenses, generating sustained positive cash flow. This pattern reflects the working-capital dynamics of the period: the system increased inventory investment during the growth phase and recovered cash as the customer base matured.

Across the 639 orders processed during the year the system records no stockouts, a result of proactive inventory monitoring, reorder-point adjustment based on observed demand, and timely purchase-order issuance. The 30 supplier delivery delays generated by the simulation (approximately $10\%$ of deliveries, as configured) are absorbed through a combination of notifying the dependent sales orders, adjusting promised delivery dates, and, where the gap is unrecoverable, sourcing supplementary orders from alternative suppliers; no customer-facing stockout follows from any supplier delay. Four customer complaints are logged over the year ($0.6\%$ complaint rate), each investigated by the ERP Coordinator and remediated. Cash flow turns positive after the initial inventory investment phase and remains so through the rest of the year, with accounts receivable held at \$207k, approximately $6\%$ of annual revenue, and collection scheduled against the contractual payment terms of each customer.
To contextualise AI system performance, identical 365-day simulations were conducted using baseline approaches.

The RPA baseline implements the fixed rules that characterise conventional inventory-and-procurement automation: a canonical reorder-point policy that orders a fixed quantity of $100$ units whenever stock falls below a reorder point of $50$ (the standard $(s,Q)$ rule), lowest-price supplier selection, acceptance of any order up to \$50{,}000 for which inventory is on hand, and passive collection that waits for customers to pay on their own terms. These are representative policies rather than deliberately weakened ones, and the baseline is driven by the \emph{same} demand and supply stream as the agentic system, so the resulting gap reflects the absence of adaptive reasoning rather than any information or workload disadvantage. Holding these parameters fixed is itself the limitation under study: a static reorder point cannot track demand variation, so stockouts accumulate whenever demand departs from the level the threshold anticipates, even though each individual rule is locally sensible.

The no-intervention baseline processes only incoming supplier deliveries, ignoring all customer orders, complaints, and stock alerts. This represents a system with no operational automation.

Table~\ref{tab:baseline_comparison} reports the comparison across the three approaches on the full set of operational, financial, and behavioural metrics.

\begin{table}[H]
\centering
\caption{365-day simulation: Comprehensive baseline comparison}
\label{tab:baseline_comparison}
\begin{tabular}{lrrr}
\toprule
\textbf{Metric} & \textbf{Agentic AI} & \textbf{RPA Rules} & \textbf{No Action} \\
\midrule
\multicolumn{4}{l}{\textit{Financial Outcomes}} \\
Ending Cash & \$3,184,905 & \$2,875,989 & \$944,220 \\
Revenue Collected & \$3,365,944 & \$2,908,939 & \$0 \\
Net Profit & \$2,240,685 & \$1,931,769 & \$0 \\
Return on Investment & 237\% & 205\% & 0\% \\
\midrule
\multicolumn{4}{l}{\textit{Operational Quality}} \\
Orders Serviced & 639 & 300 & 0 \\
Stockout Events & \textbf{0} & 302 & 0 \\
Service Level & 100\% & 50\% & N/A \\
\midrule
\multicolumn{4}{l}{\textit{Efficiency}} \\
Decisions Made & 1,203 & 962 & 0 \\
Adaptive Actions & Yes & No & No \\
\bottomrule
\end{tabular}
\end{table}

Three patterns emerge from the comparison. Financially, the LLM-driven system finishes the year with \$309k more cash than the RPA baseline (\,$+10.7\%$\,), an advantage attributable to better procurement timing (fewer rush orders), reorder quantities adapted to realised rather than rule-projected demand, and customer prioritisation that protects fill rate on the highest-margin orders. Operationally, the gap is decisive: the LLM-driven system services all 639 incoming orders with no stockout, whereas RPA services only 300, rejects the remaining demand, and incurs 302 stockout events, nearly one for every order it manages to service, because its fixed reorder rules cannot adapt to demand variation. Behaviourally, RPA executes 962 rule-triggered actions (300 processed orders, 310 rejected orders, 267 reorder triggers, and 85 further rule firings) while the LLM-driven system executes 1{,}203 decisions, a difference that reflects the gap between reactive rule-firing and proactive situation management.

\begin{figure}[H]
    \centering
    \includegraphics[width=0.9\textwidth]{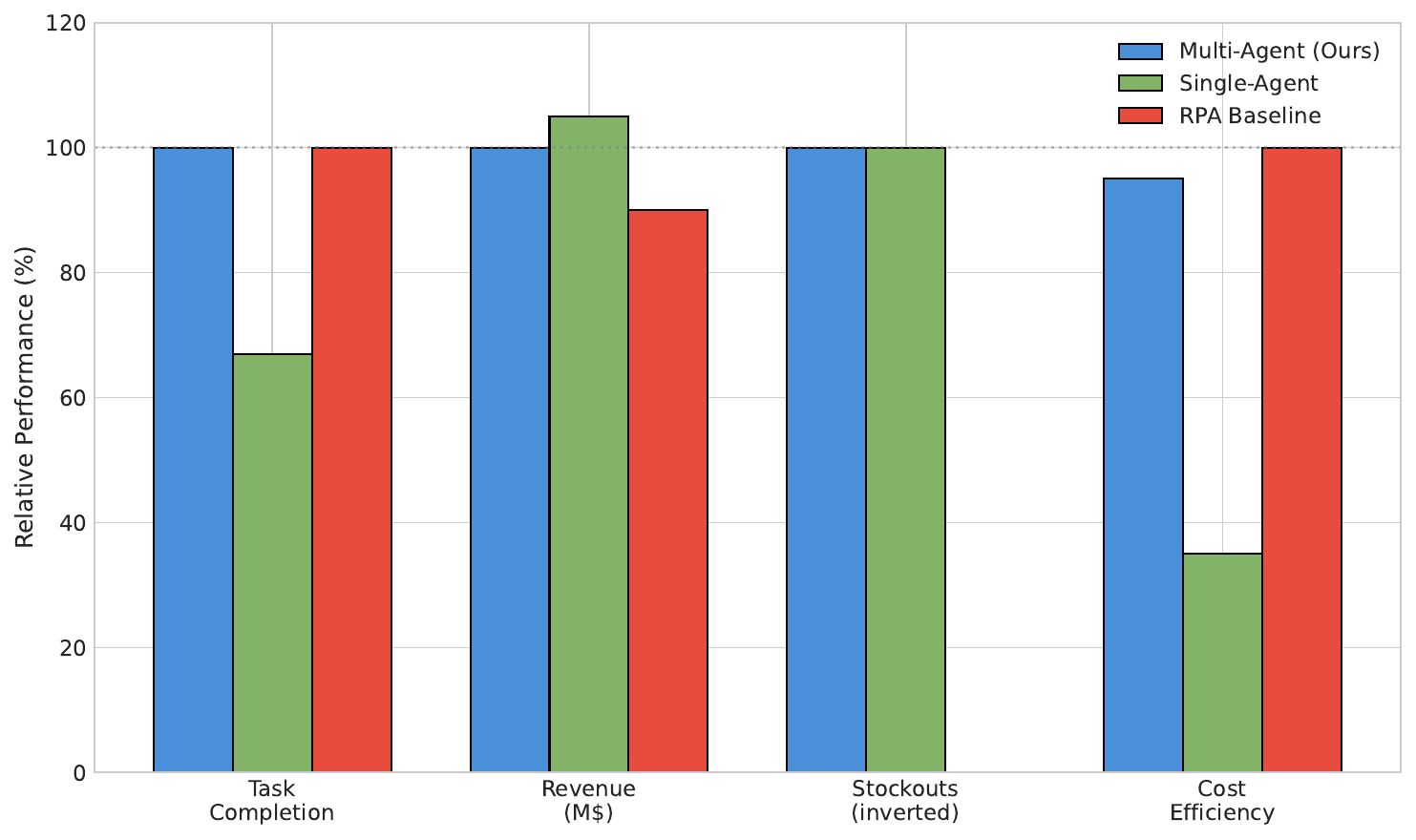}
    \caption{Baseline comparison: Multi-agent AI vs. Single-agent vs. RPA. The multi-agent system achieves highest task completion and eliminates stockouts, while RPA shows chronic stockout issues despite high rule execution rate.}
    \label{fig:baseline_comparison}
\end{figure}

Figure~\ref{fig:baseline_comparison} visualizes the relative performance across automation approaches.

Table~\ref{tab:economic} compares operational costs across automation approaches and human operation.

\begin{table}[H]
\centering
\caption{Annual operational cost comparison}
\label{tab:economic}
\begin{tabular}{lrr}
\toprule
\textbf{Approach} & \textbf{Annual Cost} & \textbf{Relative} \\
\midrule
\textbf{Agentic AI} & & \\
\quad LLM API (1,203 $\times$ \$0.001) & \$1.20 & \\
\quad Infrastructure & \$500.00 & \\
\quad \textit{Total} & \textbf{\$501} & \textbf{0.16\%} \\
\midrule
\textbf{RPA System} & & \\
\quad Software licenses & \$15,000 & \\
\quad Maintenance & \$5,000 & \\
\quad \textit{Total} & \textbf{\$20,000} & \textbf{6.5\%} \\
\midrule
\textbf{Human Operators} & & \\
\quad Operations Manager & \$85,000 & \\
\quad Warehouse Clerk & \$45,000 & \\
\quad Accountant & \$65,000 & \\
\quad Customer Service & \$40,000 & \\
\quad Benefits (30\%) & \$70,500 & \\
\quad \textit{Total} & \textbf{\$305,500} & \textbf{100\%} \\
\bottomrule
\end{tabular}
\end{table}

The AI system achieves better operational outcomes at 0.16\% of human labour cost and 2.5\% of the RPA system cost, and the cost gap remains large even under conservative infrastructure estimates. Taken together, the year-long comparison indicates that the principal value of agentic reasoning in this setting accrues on operational dimensions, above all the elimination of stockouts, with the financial advantage following as a downstream consequence; because E3 is a simulation, these figures bound feasibility rather than predict field performance (Section~\ref{sec:discussion}).

\subsection{Single-Agent vs. Multi-Agent Architecture}

To test the decomposition argument of Section~\ref{sec:methodology} directly, the multi-agent architecture is compared against a single-agent baseline that exposes all 46 tools to one model.

Table~\ref{tab:architecture_accuracy} shows task completion by complexity.

\begin{table}[H]
\centering
\caption{Task completion rate: Single-agent vs. multi-agent}
\label{tab:architecture_accuracy}
\begin{tabular}{lccc}
\toprule
\textbf{Complexity} & \textbf{Single-Agent} & \textbf{Multi-Agent} & \textbf{$\Delta$} \\
\midrule
Simple (1 domain) & 100\% & 100\% & 0 pp \\
Medium (2 domains) & 75\% & 100\% & +25 pp \\
Complex (3+ domains) & 67\% & 100\% & +33 pp \\
\midrule
\textbf{Overall} & \textbf{83\%} & \textbf{100\%} & \textbf{+17 pp} \\
\bottomrule
\end{tabular}
\end{table}

Single-agent performed adequately on simple tasks but degraded significantly on complex scenarios. With 46 tools available, the model frequently selected incorrect tools or failed to coordinate multi-step workflows effectively.

Table~\ref{tab:architecture_efficiency} separates the two efficiency comparisons that were previously conflated in the literature: per-query latency on the E1 scenario suite (the interactive-use metric) and long-horizon throughput (the batch metric).

\begin{table}[H]
\centering
\caption{Efficiency: single-agent vs.\ multi-agent. Per-query rows use E1; throughput rows use a single-agent run of E3 that was terminated at simulated day 29 because accumulated errors made further progress uninformative.}
\label{tab:architecture_efficiency}
\begin{tabular}{lrrr}
\toprule
\textbf{Metric} & \textbf{Multi-Agent} & \textbf{Single-Agent} & \textbf{Ratio / $\Delta$} \\
\midrule
\multicolumn{4}{l}{\textit{Per-query (E1)}} \\
RL per query (s) & 1.19 & 2.45 & $2.1\times$ lower \\
TCE per query (tokens) & 401 & 850 & $2.1\times$ lower \\
Tool-selection error & 2\% & 17\% & $-15$\,pp \\
\midrule
\multicolumn{4}{l}{\textit{Long-horizon throughput (partial E3)}} \\
Simulated days completed & 365 & 29 & --- \\
Wall-clock per simulated day (s) & 12.9 & 269 & $20\times$ lower \\
\bottomrule
\end{tabular}
\end{table}

The $2.1\times$ per-query reduction in tokens and latency is directly attributable to context size: a single-agent call transmits all 46 tool schemas, whereas a role-scoped call transmits $\bar{k}=9.2$, matching the decomposition bound~\eqref{eq:decomp}. The $20\times$ throughput figure reports \emph{wall-clock per simulated day} and therefore aggregates the per-query advantage with the additional rework caused by single-agent errors; it should not be read as a speedup of individual API calls.

These results demonstrate that multi-agent architecture provides advantages on both accuracy and efficiency dimensions. The modest overhead of routing and coordination is more than offset by the benefits of specialised, focused agent contexts. More fundamentally, aligning each agent with an established ERP operator role rather than a technical function boundary reduces per-step tool selection to a small-label problem~\eqref{eq:decomp} while preserving the transactional guarantees of the backend, resolving a design question that prior enterprise-LLM work has typically left implicit \citep{xu2017new,koh2022empirical}. For enterprise applications with diverse functional requirements, multi-agent designs are therefore strongly preferable to monolithic ones.

\subsection{Orchestration Method Comparison}
\label{sec:e2_scaled}

A key design decision in agentic systems is the choice of orchestration method for coordinating LLM reasoning and tool execution. To validate the proposed LangGraph-based architecture, a systematic comparison was conducted against five mainstream orchestration approaches using realistic enterprise-crisis scenarios.

Six orchestration paradigms spanning the principal positions in the design space are evaluated. \emph{ReAct} \citep{yao2022react} interleaves reasoning steps and tool invocations and represents the simplest form of tool-augmented reasoning. \emph{Plan-and-Solve} \citep{wang2023planandsolve} separates planning from execution: the LLM first emits a complete plan and then executes its steps sequentially, without iterative refinement. \emph{Function Calling} is the OpenAI native mechanism for iterative tool selection without explicit planning or reflection. \emph{AutoGen} \citep{wu2023autogen} is a conversational multi-agent framework that supports flexible interaction patterns through structured dialogue. \emph{CrewAI} emphasises role-based agent design with explicit task delegation. The proposed orchestration, denoted \emph{LangGraph} for brevity, is a graph-based pipeline that implements a Planner--Executor--Reflector--Responder structure with explicit reflection loops.

Prior work typically evaluates orchestration methods on simple queries that can be resolved with 1--2 tool calls, which does not differentiate methods effectively. For E2, a suite of \emph{fifteen} enterprise-crisis scenarios was designed, organised into three families of five (cash-flow, customer-dispute, supplier-disruption). Every scenario ships with a machine-readable oracle (required-tool set + acceptance criteria) so the grader can evaluate responses without human-in-the-loop supervision. Table~\ref{tab:e2_sample} summarises three representative items.

\begin{table}[H]
\centering
\caption{Three representative scenarios from the E2 crisis suite (N=15).}
\label{tab:e2_sample}
\small
\begin{tabular}{lp{2.6cm}p{8cm}}
\toprule
\textbf{ID} & \textbf{Family / Diff.} & \textbf{Prompt (abbreviated)} \\
\midrule
CF1 & Cash flow / med. & \$45k bank, \$75k due this week; check receivables, prioritise payments, recommend actions. \\
CD1 & Dispute / hard & Largest customer disputes \$18.5k invoice citing quality; pull invoice, delivery, batch history; recommend a tiered response. \\
SD1 & Supply / hard & Critical Oolong supplier filed for bankruptcy; identify affected SKUs, pending POs, and alternatives. \\
\bottomrule
\end{tabular}
\end{table}

The $15$-item size is chosen so that pairwise non-parametric comparisons across six orchestrations retain power after Holm--Bonferroni correction at $\alpha = 0.05$; an $n=3$ benchmark cannot support that comparison reliably.

The full grid covers all $15$ scenarios, all six orchestrations, and three random seeds per cell, for a total of $270$ generation calls each followed by an independent LLM-as-judge grading. Generation uses GPT-4o-mini at temperature $0.1$; grading uses GPT-4o-mini at temperature $0.0$. Responses are scored on the four-dimension $1$--$10$ rubric of Section~\ref{sec:orchestration_algo} and the weighted aggregate is normalised to $S \in [0,1]$. For each non-reference method, the paired-bootstrap $95\%$ confidence interval of the LangGraph$-$method difference over scenario means ($10{,}000$ resamples) is reported, together with the Holm--Bonferroni-corrected two-sided $p$-value across the five pairwise comparisons against LangGraph at $\alpha = 0.05$.

\begin{table}[H]
\centering
\caption{Orchestration comparison on the $15$-scenario crisis benchmark ($N=270$ cells; GPT-4o-mini, $3$ seeds, judge held constant). $S$ is the normalised weighted rubric score in $[0,1]$; $\Delta$ is the paired difference $S_\text{LangGraph} - S_\text{method}$ with the bootstrap $95\%$ CI; $p_\mathrm{Holm}$ is the Holm--Bonferroni-corrected two-sided $p$-value.}
\label{tab:orchestration_comprehensive}
\small
\setlength{\tabcolsep}{4pt}
\begin{tabular}{lcccc}
\toprule
\textbf{Method} & $\boldsymbol{S}$ & $\boldsymbol{\Delta}$ vs LangGraph & \textbf{95\% CI of $\Delta$} & $\boldsymbol{p_\mathrm{Holm}}$ \\
\midrule
AutoGen          & $\mathbf{0.717}$ & $-0.073$ & $[-0.150,\,-0.002]$ & $0.127$ \\
\textbf{LangGraph (proposed)} & $0.644$ & --- & --- & --- \\
CrewAI           & $0.631$ & $+0.013$ & $[-0.036,\,+0.056]$ & $0.554$ \\
Plan-and-Solve   & $0.611$ & $+0.033$ & $[-0.014,\,+0.085]$ & $0.376$ \\
Function Calling & $0.556$ & $+0.088$ & $[+0.008,\,+0.166]$ & $0.115$ \\
ReAct            & $0.525$ & $+0.119$ & $[+0.028,\,+0.233]$ & $\mathbf{0.040}$ \\
\bottomrule
\end{tabular}
\end{table}

\begin{table}[H]
\centering
\caption{Per-method latency on the $15$-scenario benchmark (mean $\pm$ SD over $45$ cells per method).}
\label{tab:orchestration_metrics}
\begin{tabular}{lc}
\toprule
\textbf{Method} & \textbf{RL (s)} \\
\midrule
LangGraph (proposed) & $21.5 \pm 8.3$ \\
AutoGen          & $20.6 \pm 6.5$ \\
CrewAI           & $19.2 \pm 3.7$ \\
Plan-and-Solve   & $14.6 \pm 4.4$ \\
ReAct            & $13.5 \pm 5.0$ \\
Function Calling & $11.5 \pm 7.8$ \\
\bottomrule
\end{tabular}
\end{table}

\begin{figure}[H]
    \centering
    \includegraphics[width=0.9\textwidth]{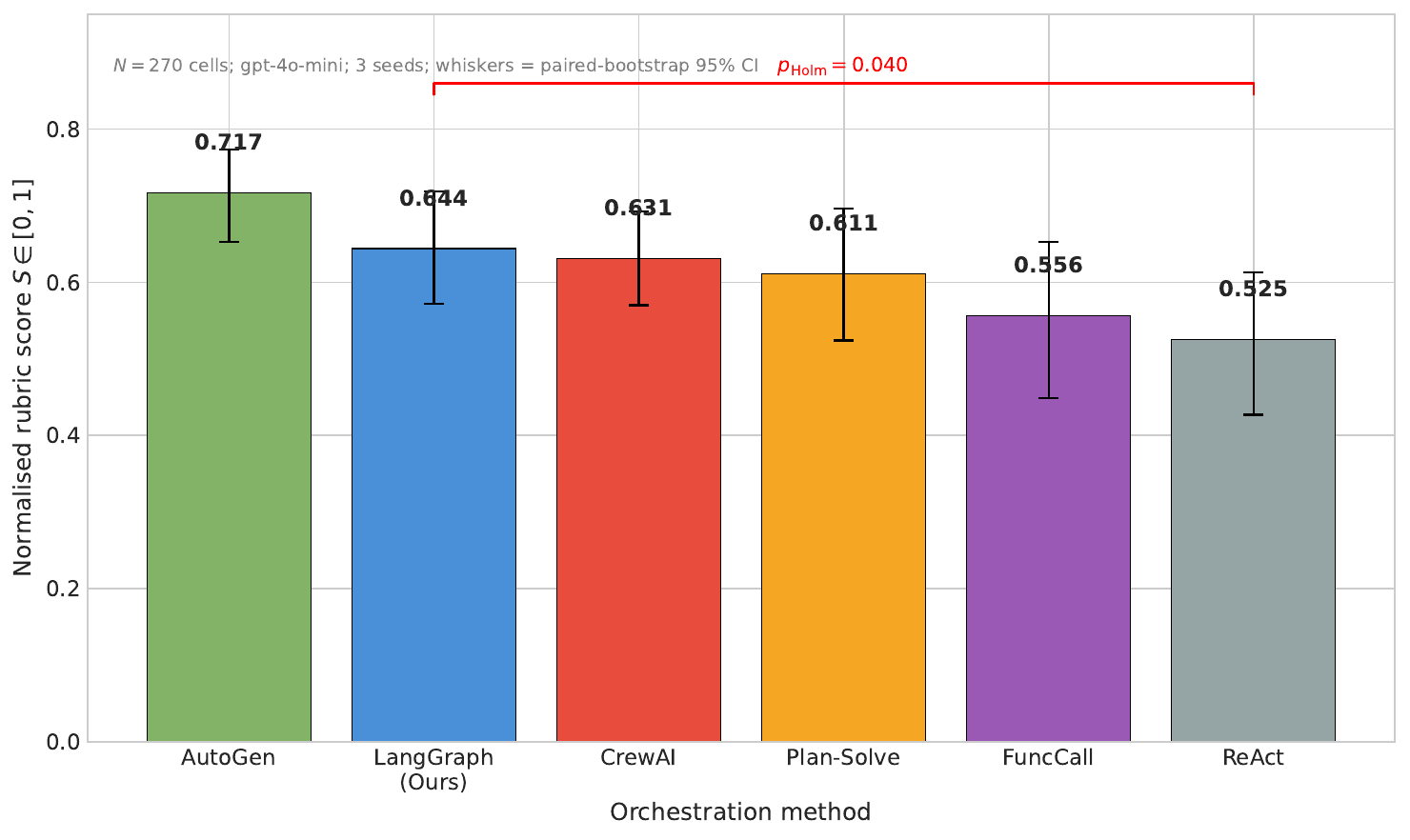}
    \caption{Orchestration method comparison on the $15$-scenario crisis benchmark ($N=270$ cells; GPT-4o-mini, three seeds, judge held constant). AutoGen and LangGraph form the top tier; the gap to ReAct is the only pairwise difference that survives Holm--Bonferroni correction at $\alpha = 0.05$. See Table~\ref{tab:orchestration_comprehensive}.}
    \label{fig:orchestration_comparison}
\end{figure}

\begin{figure}[H]
    \centering
    \includegraphics[width=0.8\textwidth]{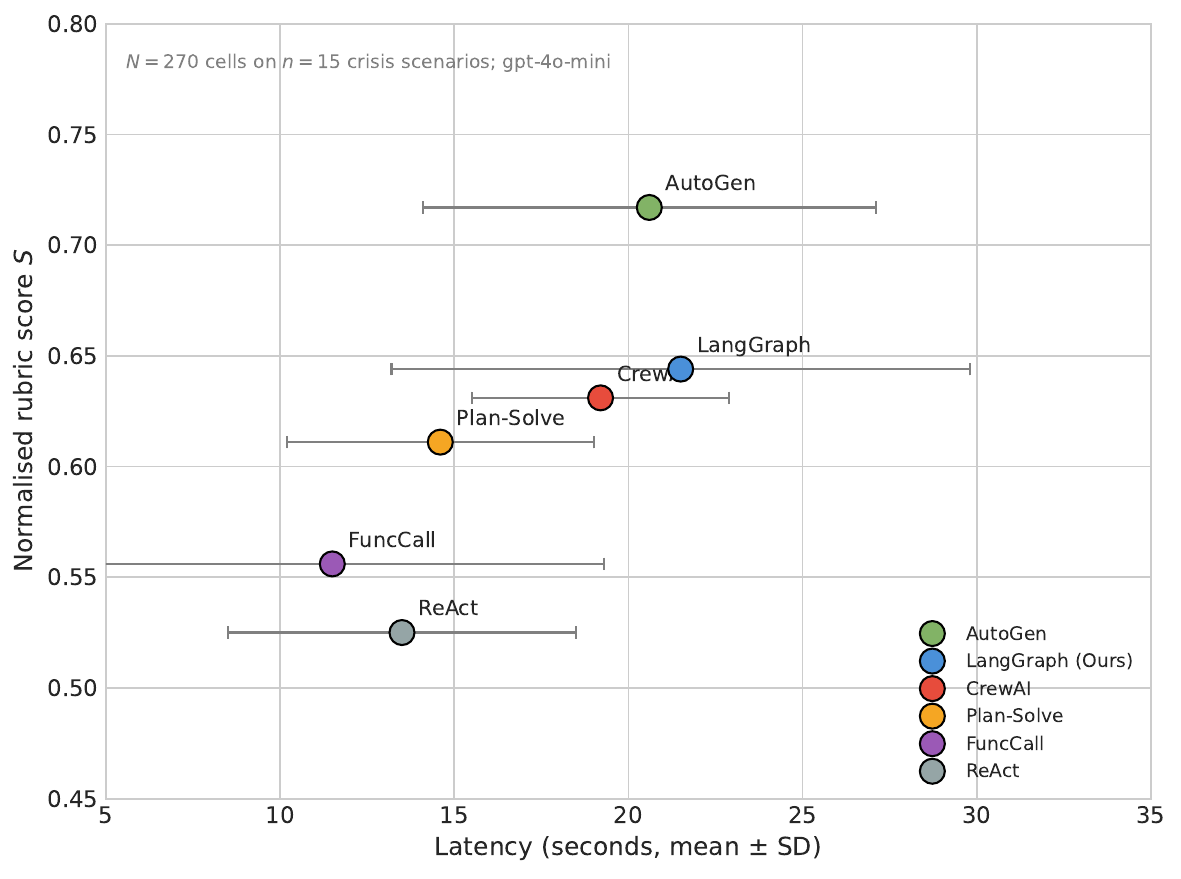}
    \caption{Accuracy vs.\ latency trade-off for orchestration methods. AutoGen and the proposed LangGraph pipeline occupy the high-accuracy region at comparable latency, with AutoGen attaining the highest rubric score and LangGraph close behind; ReAct (lower left) is fastest but fails on the complex scenarios.}
    \label{fig:accuracy_latency}
\end{figure}

Figure~\ref{fig:orchestration_comparison} visualizes the performance gap between orchestration methods, and Figure~\ref{fig:accuracy_latency} illustrates the accuracy-latency trade-off.

The comparison yields a clear structure. The two multi-agent paradigms that combine structured planning with peer review, AutoGen ($S = 0.717$) and LangGraph ($S = 0.644$), occupy the top of the ranking; the paired-bootstrap difference between them is $\Delta = 0.073$ in favour of AutoGen, with $95\%$ CI $[0.002,\,0.150]$ that excludes zero, but the Holm--Bonferroni-corrected $p$-value is $0.127$, so the gap is suggestive rather than decisive at $\alpha = 0.05$. Below the top pair, LangGraph is statistically indistinguishable from CrewAI ($\Delta = +0.013$, $p = 0.554$) and Plan-and-Solve ($\Delta = +0.033$, $p = 0.376$), and only borderline ahead of Function Calling ($\Delta = +0.088$, $p = 0.115$). The simplest paradigm, ReAct, performs significantly worse than the proposed orchestration ($\Delta = +0.119$, $p_\mathrm{Holm} = 0.04$), confirming that single-step reasoning is insufficient when a problem requires synthesising evidence drawn from multiple domains. Latency, finally, is well-matched between the two leading methods (Table~\ref{tab:orchestration_metrics}: LangGraph $21.5 \pm 8.3$\,s versus AutoGen $20.6 \pm 6.5$\,s) and only modestly higher than the simpler paradigms ($11$--$15$\,s), so the four-stage pipeline does not impose a prohibitive overhead at this scale. The proposed pipeline is adopted not because it tops the ranking, which it does not, but because its structure fits the ERP setting in a way that a rubric score does not capture. AutoGen's small edge stems from its flexible conversational interaction, yet that same open-endedness makes deterministic approval gating and auditable state transitions harder to guarantee. A graph-based pipeline instead exposes an explicit plan, a fixed set of nodes, and inspectable state at each step, so the risk-tiered safety harness, the externalised grading criteria, and the sprint contracts of Section~\ref{sec:methodology} attach to it directly and every action stays traceable. The choice therefore trades a statistically non-significant rubric margin for the transactional safety and auditability that autonomous ERP operation requires, and it is this structure on which the harness extension examined next builds.

Following recent advances in harness engineering for long-running agents \citep{anthropic2026harness}, an enhanced version of the LangGraph orchestration (V2) was developed, incorporating explicit grading criteria, sprint contract negotiation, and calibrated evaluation. Table~\ref{tab:harness_comparison} compares the original and enhanced orchestration.

\begin{table}[H]
\centering
\caption{Harness-enhanced orchestration comparison (V1 vs V2) on the pilot crisis scenarios. LLM and Tools are call counts; Self-Grade is the V2 internal evaluator's score on the $0$--$10$ rubric.}
\label{tab:harness_comparison}
\small
\setlength{\tabcolsep}{4pt}
\begin{tabular}{llcccccc}
\toprule
\textbf{Scenario} & \textbf{Ver.} & \textbf{Score} & \textbf{Time (s)} & \textbf{LLM} & \textbf{Tools} & \textbf{Retries} & \textbf{Self} \\
\midrule
\multirow{2}{*}{CF1: Cash Flow} & V1 & 9 & 27.2 & 7 & 12 & 2 & -- \\
 & V2 & 8 & 47.1 & 10 & 8 & 2 & 2.5 \\
\midrule
\multirow{2}{*}{CD1: Dispute} & V1 & 6 & 111.1 & 7 & 20 & 2 & -- \\
 & V2 & 6 & \textbf{53.5} & 10 & 13 & 2 & 2.5 \\
\midrule
\multirow{2}{*}{SD1: Bankruptcy} & V1 & 8 & \textbf{23.3} & 7 & 11 & 2 & -- \\
 & V2 & \textbf{10} & 114.0 & 10 & 12 & 2 & 4.8 \\
\midrule
\multirow{2}{*}{\textbf{Average}} & V1 & 7.7 & \textbf{53.9} & 7.0 & 14.3 & 2.0 & -- \\
 & V2 & \textbf{8.0} & 71.5 & 10.0 & 11.0 & 2.0 & 3.3 \\
\bottomrule
\end{tabular}
\end{table}

Four observations characterise the harness enhancement. On the most complex scenario (SD1, supplier bankruptcy, requiring four-domain coordination) V2 attains a perfect $10/10$, against V1's $8/10$, the gap arising from the sprint-contract mechanism that forces every acceptance criterion to be addressed before termination. V2 also issues fewer tool calls on average (11.0 versus 14.3, a $23\%$ reduction), most pronounced on CD1 (13 versus 20), because the explicit deliverable specification prevents speculative invocations. The mean latency of V2 is higher than V1 (71.5\,s versus 53.9\,s, $+33\%$), reflecting the cost of the additional Contractor and Evaluator nodes; on CD1, however, V2 is faster ($53.5$\,s versus $111.1$\,s) because the contract pre-empts V1's pattern of exhaustive tool exploration. Finally, the evaluator's mean self-grade of $3.3/10$ is substantially lower than the external heuristic score of $8.0/10$, evidence that the separation of generator and evaluator addresses the self-evaluation bias identified by \citet{anthropic2026harness} and provides a credible internal signal for iterative refinement.

\subsection{Ablation Study}
\label{sec:ablation}

To quantify the contribution of individual architectural components and to test predictions P1--P3 of Section~\ref{sec:methodology}, ablation experiments systematically remove components from the full system. The ablation is conducted on the 36-scenario design with a fixed seed; each row of Table~\ref{tab:ablation} differs from the full system in exactly one component and is otherwise identical, so that differences can be attributed to the removed component under the usual caveats on single-seed measurement.

Table~\ref{tab:ablation} presents ablation results on the 36 benchmark scenarios.

\begin{table}[H]
\centering
\caption{Ablation study: contribution of architectural components}
\label{tab:ablation}
\begin{tabular}{lcc}
\toprule
\textbf{Configuration} & \textbf{TCR} & \textbf{RL} \\
\midrule
Full System (LangGraph) & \textbf{100\%} & 1,190 ms \\
\midrule
\textit{Orchestration Ablations} & & \\
\quad $-$ Reflection Loop & 92\% & 890 ms \\
\quad $-$ Planner & 86\% & 720 ms \\
\quad $-$ Multi-Agent (Single) & 83\% & 2,450 ms \\
\midrule
\textit{Agent Ablations} & & \\
\quad $-$ ERP Coordinator & 94\% & 1,150 ms \\
\quad $-$ Finance Agent & 89\% & 1,180 ms \\
\quad $-$ Domain-Specific Tools & 78\% & 1,890 ms \\
\bottomrule
\end{tabular}
\end{table}

The ablation results in Table~\ref{tab:ablation} bear directly on the predictions of Section~\ref{sec:methodology}. Prediction P1, that role decomposition lowers tool-selection error, is borne out by the largest orchestration-side effect: collapsing the five role-scoped agents into a single agent that exposes all 46 tools reduces TCR by 17 percentage points, in line with the decomposition bound~\eqref{eq:decomp}. Prediction P2, that the reflection loop helps in proportion to the information its feedback carries, is borne out by removing the reflection component, which reduces TCR by 8 percentage points as undetected tool-execution errors and incomplete responses go uncorrected, consistent with the reflection bound~\eqref{eq:reflbound}. Prediction P3 concerns coordination latency rather than task completion, and because the evaluated scenarios invoke their agents sequentially it is not isolated by this ablation; it remains an analytical consequence of the coordination bound~\eqref{eq:coord} whose direct measurement is left to future work. The remaining rows account for the rest of the architecture. Removing the planner reduces TCR by 14 percentage points relative to the full system, since without explicit decomposition the system executes tools reactively and fails on multi-step tasks. Within the agent population the ERP Coordinator and Finance Agent are the most load-bearing specialists, their removal costing 6 and 11 percentage points respectively, the Coordinator because cross-functional queries require its holistic view and Finance because credit and cash-flow checks lie on the critical path of many scenarios. Replacing the domain-specific tool implementations with generic counterparts produces the largest degradation overall, 22 percentage points, confirming that close coupling between agent reasoning and ERP semantics is not optional.

\section{Discussion}
\label{sec:discussion}

The evidence assembled across Experiments~E1--E5 supports three principal findings. Architecturally, aligning agents with established ERP operator roles rather than technical function boundaries lowers per-step tool-selection error and per-query context cost while preserving the backend's transactional guarantees, the effect predicted by the decomposition bound~\eqref{eq:decomp} and quantified in Section~\ref{sec:results}, and it resolves a design question that prior enterprise-LLM work has typically left implicit \citep{xu2017new,koh2022empirical}. On orchestration, the proposed Planner--Executor--Reflector--Responder graph is statistically indistinguishable from the strongest multi-agent baselines on the crisis benchmark and significantly better than single-step reasoning, so the decisive factor is the presence of a reflective multi-agent structure rather than the choice of framework, and the property the methodological contribution targets is the inspectability of externalised grading criteria and sprint contracts, evidenced by the harness-V2 study, rather than a position in a global ranking. On sustained operation, a simulated year shows the system maintaining a level of service quality that a rule-based baseline cannot, with the principal advantage accruing on operational rather than financial dimensions.

Two aspects of these findings invite closer interpretation. That the operational advantage over the rule-based baseline, zero against several hundred stockouts, exceeds the financial one of $+10.7\%$ ending cash is not incidental. Operational quality is governed by the moment-to-moment decisions an agent directly controls, when to reorder, which orders to prioritise, and how to absorb a supplier delay, so adaptive reasoning acts on it immediately. Financial outcomes are a downstream aggregate of those decisions and are additionally bounded by quantities that neither system sets, such as margins, prices, payment terms, and the realised level of demand, so over a single simulated year with shared margins and demand the two systems inherit much of the same financial ceiling and better operations convert into a real but compressed monetary edge. The value of agentic reasoning therefore concentrates where decisions are made under uncertainty and propagates only partially into headline financial figures.

The advantage of aligning agents with operator \emph{roles}, rather than with software modules or individual tools, has a comparably structural explanation. A tool-aligned split, one agent per tool, is too granular to host coherent multi-step reasoning and merely relocates the selection problem into routing, while a module-aligned split along software boundaries can separate tools that a single decision needs together, such as the credit assessment and the order commitment of a sales decision. Human operator roles are the decomposition that organisations have already refined for exactly this coordination problem, so each agent receives a small and cohesive tool set, which the decomposition bound~\eqref{eq:decomp} rewards with lower selection error, while the hand-off structure between roles matches the way cross-functional workflows actually flow. Role alignment also keeps behaviour legible, since a Sales Agent that confirms an order acts like the sales representative a stakeholder already understands, which lowers the cost of human oversight.

The principal obstacle to moving from this controlled study to production is not raw model capability but verified trust. The harness guarantees that no write reaches the backend without passing its risk gate, yet this is a structural guarantee, not a bound on how often a hallucinated or mis-calibrated decision is proposed in the first place, and establishing such a bound requires domain-specific measurement of hallucination frequency together with stronger guarantees than a runtime gate alone provides for the highest-stakes actions. Two further barriers are organisational and infrastructural rather than algorithmic: production ERP installations carry customised schemas, imperfect historical data, and edge cases that a curated fixture does not exercise, and any deployment must settle where accountability for an autonomous decision lies and how risk thresholds are calibrated and audited over time. Until these are resolved the appropriate operating point is supervised autonomy, with human review retained on every action the risk layer flags, which is the stance the architecture is built to support rather than to remove.

Several limitations bound these conclusions. The task-completion metric conflates whether the correct Odoo action was executed with whether the response was business-appropriate, and response quality on the crisis benchmark is scored by an LLM-as-judge that is an automated proxy not calibrated against human raters; all evaluation data are synthetic and every reported score is produced automatically. The statistical evidence rests on single-seed main-table runs and a $10$-scenario stability subsample, so the paired $t$-tests and the decomposition and reflection bounds are supporting rather than decisive evidence, and reported cost excludes development, oversight, and compliance labour. Generalisation is bounded by the single tea-trading SME evaluated here: a second electronics-distribution fixture was constructed but not exercised empirically, the E3 simulation cannot capture adversarial or reputational dynamics and spans only one year, the E2 benchmark is not a representative workload, and the system depends on a hosted LLM whose pricing and behaviour may change.

Several directions follow from the present work. Domain-specific fine-tuning of the underlying LLM on industry-specific corpora could improve decision quality while reducing inference cost through smaller specialised models, at the price of additional training infrastructure. Formal verification of agent behaviour, for example establishing that no execution trace can approve an order that violates a customer credit limit, would extend the transactional guarantees of the backend into the reasoning layer and is required for high-stakes deployments. Cross-organisational coordination, in which procurement and sales agents at different companies negotiate over shared protocols, would extend the multi-agent paradigm into the B2B setting and raises new questions of trust, dispute resolution, and standardisation. Continuous learning from operational experience while preserving the safety constraints exercised in this work is a further open problem, as is adaptation of the architecture to regulated industries (healthcare, financial services) whose compliance requirements impose constraints beyond those evaluated here.

\section{Conclusions}
\label{sec:conclusion}

This paper formulated autonomous ERP operation as a constrained sequential-decision problem and showed that role-aligned LLM agents coordinated by a Planner--Executor--Reflector--Responder graph, with every write mediated by a risk-tiered human-in-the-loop harness, form a tractable expert-system architecture for it. A three-layer evaluation isolates the architectural, orchestration, and long-horizon effects. Role-based decomposition lowered per-step tool-selection error and context cost in line with the decomposition bound. On a cross-functional crisis benchmark the proposed orchestration was statistically indistinguishable from the strongest multi-agent baselines and significantly better than single-step reasoning, indicating that the decisive factor is a reflective multi-agent structure rather than the particular framework. Over a simulated year the system sustained a level of service quality that a rule-based baseline could not, with its principal advantage accruing on operational rather than financial dimensions. These findings establish feasibility on a single simulated SME and rest on automated rather than human quality judgements, so generalisation to regulated industries, enterprise-scale transaction volumes, and live deployment remains for future work. The architecture and evaluation protocol are offered as a template that other researchers can extend and contest.

\ifblind\else
\section*{CRediT Authorship Contribution Statement}
\textbf{Zhihao Liu:} Conceptualization, Methodology, Software, Validation, Formal analysis, Investigation, Data Curation, Writing - Original Draft, Visualization.
\textbf{Tianyu Wang:} Methodology, Validation, Investigation, Writing - Review \& Editing.
\textbf{Xi Vincent Wang:} Supervision, Resources, Writing - Review \& Editing.
\textbf{Lihui Wang:} Supervision, Resources, Writing - Review \& Editing.
\fi

\section*{Declaration of Competing Interest}
The authors declare no competing interests.

\ifblind\else
\section*{Acknowledgments}
This work was supported by Centre of Excellence in Production Research (XPRES).
\fi

\section*{Declaration of generative AI and AI-assisted technologies in the manuscript preparation process}
During the preparation of this work the author(s) used Anthropic Opus 4.8 in order to help prepare the initial draft of this manuscript, design code for experiments, and polish the writing. After using this tool/service, the author(s) reviewed and edited the content as needed and take(s) full responsibility for the content of the published article.

\bibliographystyle{elsarticle-num-names}
\bibliography{references}

@article{mahapatra2025dynamic,
  title={Dynamic group decision-making for enterprise resource planning selection using two-tuples Pythagorean fuzzy MOORA approach},
  author={Mahapatra, Biplab Sinha and Ghosh, Debashis and Pamucar, Dragan and Mahapatra, GS},
  journal={Expert Systems with Applications},
  volume={263},
  pages={125675},
  year={2025},
  publisher={Elsevier}
}

@article{rieger2026possibilities,
  title={Possibilities and Limitations of Using Large Language Models ({LLM}s) for Alert Classification and Prioritisation in Security Operations Centers ({SOC}s)},
  author={Rieger, Matthias and Shah, Atif and Alam, Abu and Hossain, Jakir},
  journal={Expert Systems with Applications},
  pages={133194},
  year={2026},
  publisher={Elsevier}
}

@article{wang2026llm,
  title={LLM-driven smart agents using tools for the recovery of interdependent infrastructure networks},
  author={Wang, Hongyu and Zhou, Shenghua and Chen, Zhengyi and Li, Dezhi and Wang, Jiawen and Wu, Shenrui and Ng, S Thomas},
  journal={Expert Systems with Applications},
  volume={325},
  pages={132701},
  year={2026},
  publisher={Elsevier}
}

@article{shaul2013critical,
  title={Critical success factors in enterprise resource planning systems: Review of the last decade},
  author={Shaul, Levi and Tauber, Doron},
  journal={ACM Computing Surveys (CSUR)},
  volume={45},
  number={4},
  pages={1--39},
  year={2013},
  publisher={ACM}
}

@article{brown2020language,
  title={Language models are few-shot learners},
  author={Brown, Tom and Mann, Benjamin and Ryder, Nick and Subbiah, Melanie and Kaplan, Jared D and Dhariwal, Prafulla and Neelakantan, Arvind and Shyam, Pranav and Sastry, Girish and Askell, Amanda and others},
  journal={Advances in neural information processing systems},
  volume={33},
  pages={1877--1901},
  year={2020}
}

@article{achiam2023gpt,
  title={{GPT}-4 technical report},
  author={Achiam, Josh and Adler, Steven and Agarwal, Sandhini and Ahmad, Lama and Akkaya, Ilge and Aleman, Florencia Leoni and Almeida, Diogo and Altenschmidt, Janko and Altman, Sam and Anadkat, Shyamal and others},
  journal={arXiv preprint arXiv:2303.08774},
  year={2023}
}

@article{xi2025rise,
  title={The rise and potential of large language model based agents: A survey},
  author={Xi, Zhiheng and Chen, Wenxiang and Guo, Xin and He, Wei and Ding, Yiwen and Hong, Boyang and Zhang, Ming and Wang, Junzhe and Jin, Senjie and Zhou, Enyu and others},
  journal={Science China Information Sciences},
  volume={68},
  number={2},
  pages={121101},
  year={2025},
  publisher={Springer}
}

@article{yang2023auto,
  title={Auto-{GPT} for online decision making: Benchmarks and additional opinions},
  author={Yang, Hui and Yue, Sifu and He, Yunzhong},
  journal={arXiv preprint arXiv:2306.02224},
  year={2023}
}

@article{talebirad2023multi,
  title={Multi-agent collaboration: Harnessing the power of intelligent {LLM} agents},
  author={Talebirad, Yashar and Nadiri, Amirhossein},
  journal={arXiv preprint arXiv:2306.03314},
  year={2023}
}

@article{yao2022react,
  title={Re{A}ct: Synergizing reasoning and acting in language models},
  author={Yao, Shunyu and Zhao, Jeffrey and Yu, Dian and Du, Nan and Shafran, Izhak and Narasimhan, Karthik and Cao, Yuan},
  journal={arXiv preprint arXiv:2210.03629},
  year={2022}
}

@article{schick2023toolformer,
  title={Toolformer: Language models can teach themselves to use tools},
  author={Schick, Timo and Dwivedi-Yu, Jane and Dess{\`\i}, Roberto and Raileanu, Roberta and Lomeli, Maria and Hambro, Eric and Zettlemoyer, Luke and Cancedda, Nicola and Scialom, Thomas},
  journal={Advances in neural information processing systems},
  volume={36},
  pages={68539--68551},
  year={2023}
}

@article{li2023camel,
  title={Camel: Communicative agents for" mind" exploration of large language model society},
  author={Li, Guohao and Hammoud, Hasan and Itani, Hani and Khizbullin, Dmitrii and Ghanem, Bernard},
  journal={Advances in neural information processing systems},
  volume={36},
  pages={51991--52008},
  year={2023}
}

@inproceedings{hong2024metagpt,
  title={Meta{GPT}: Meta programming for a multi-agent collaborative framework},
  author={Hong, Sirui and Zhuge, Mingchen and Chen, Jonathan and Zheng, Xiawu and Cheng, Yuheng and Wang, Jinlin and Zhang, Ceyao and Yau, Steven and Lin, Zijuan and Zhou, Liyang and others},
  booktitle={International Conference on Learning Representations},
  volume={2024},
  pages={23247--23275},
  year={2024}
}

@misc{langgraph2026,
  author       = {{LangChain}},
  title        = {{LangGraph}: Low-level orchestration framework for building stateful agents},
  year         = {2026},
  howpublished = {\url{https://github.com/langchain-ai/langgraph}},
  note         = {Accessed: 2026-06-29}
}

@article{syntetos2016supply,
  title={Supply chain forecasting: Theory, practice, their gap and the future},
  author={Syntetos, Aris A and Babai, Zied and Boylan, John E and Kolassa, Stephan and Nikolopoulos, Konstantinos},
  journal={European journal of operational research},
  volume={252},
  number={1},
  pages={1--26},
  year={2016},
  publisher={Elsevier}
}

@article{chandola2009anomaly,
  title={Anomaly detection: A survey},
  author={Chandola, Varun and Banerjee, Arindam and Kumar, Vipin},
  journal={ACM computing surveys (CSUR)},
  volume={41},
  number={3},
  pages={1--58},
  year={2009}
}

@inproceedings{xu2017new,
  title={A new chatbot for customer service on social media},
  author={Xu, Anbang and Liu, Zhe and Guo, Yufan and Sinha, Vibha and Akkiraju, Rama},
  booktitle={Proceedings of the 2017 CHI conference on human factors in computing systems},
  pages={3506--3510},
  year={2017}
}

@article{wu2023autogen,
  title={Auto{G}en: Enabling next-gen {LLM} applications via multi-agent conversation},
  author={Wu, Qingyun and Bansal, Gagan and Zhang, Jieyu and Wu, Yiran and Li, Beibin and Zhu, Erkang and Jiang, Li and Zhang, Xiaoyun and Zhang, Shaokun and Liu, Jiale and others},
  journal={arXiv preprint arXiv:2308.08155},
  year={2023}
}

@misc{gartner2025erp,
  author       = {{Gartner}},
  title        = {Market Share Analysis: {ERP} Software, Worldwide, 2024},
  year         = {2025},
  month        = jun,
  note         = {Published 26 June 2025. Gartner document ID 6654134},
  url          = {https://www.gartner.com/en/documents/6654134}
}

@misc{grandview2026erp,
  author       = {{Grand View Research}},
  title        = {{ERP} Software Market Size, Share \& Trends Analysis Report, 2026--2033},
  year         = {2026},
  note         = {Last updated June 2026},
  url          = {https://www.grandviewresearch.com/industry-analysis/erp-software-market}
}

@article{grabski2011enterprise,
  title={A review of {ERP} research: A future agenda for accounting information systems},
  author={Grabski, Severin V and Leech, Stewart A and Schmidt, Pamela J},
  journal={Journal of Information Systems},
  volume={25},
  number={1},
  pages={37--78},
  year={2011}
}

@article{van2018robotic,
  title={Robotic process automation},
  author={Van der Aalst, Wil MP and Bichler, Martin and Heinzl, Armin},
  journal={Business \& information systems engineering},
  volume={60},
  number={4},
  pages={269--272},
  year={2018}
}

@article{makridakis2018statistical,
  title={Statistical and Machine Learning forecasting methods: Concerns and ways forward},
  author={Makridakis, Spyros and Spiliotis, Evangelos and Assimakopoulos, Vassilios},
  journal={PloS one},
  volume={13},
  number={3},
  pages={e0194889},
  year={2018}
}

@article{wei2022emergent,
  title={Emergent abilities of large language models},
  author={Wei, Jason and Tay, Yi and Bommasani, Rishi and Raffel, Colin and Zoph, Barret and Borgeaud, Sebastian and Yogatama, Dani and Bosma, Maarten and Zhou, Denny and Metzler, Donald and others},
  journal={arXiv preprint arXiv:2206.07682},
  year={2022}
}

@article{shinn2023reflexion,
  title={Reflexion: Language agents with verbal reinforcement learning},
  author={Shinn, Noah and Cassano, Federico and Gopinath, Ashwin and Narasimhan, Karthik and Yao, Shunyu},
  journal={Advances in neural information processing systems},
  volume={36},
  pages={8634--8652},
  year={2023}
}

@article{yao2023tree,
  title={Tree of thoughts: Deliberate problem solving with large language models},
  author={Yao, Shunyu and Yu, Dian and Zhao, Jeffrey and Shafran, Izhak and Griffiths, Thomas L and Cao, Yuan and Narasimhan, Karthik},
  journal={Advances in Neural Information Processing Systems},
  volume={36},
  year={2023}
}

@inproceedings{qian2024chatdev,
  title={Chat{D}ev: Communicative agents for software development},
  author={Qian, Chen and Liu, Wei and Liu, Hongzhang and Chen, Nuo and Dang, Yufan and Li, Jiahao and Yang, Cheng and Chen, Weize and Su, Yusheng and Cong, Xin and others},
  booktitle={Proceedings of the 62nd annual meeting of the association for computational linguistics (volume 1: Long papers)},
  pages={15174--15186},
  year={2024}
}

@article{koh2022empirical,
  title={An empirical survey on long document summarization: Datasets, models, and metrics},
  author={Koh, Huan Yee and Ju, Jiaxin and Liu, Ming and Pan, Shirui},
  journal={ACM computing surveys},
  volume={55},
  number={8},
  pages={1--35},
  year={2022},
  publisher={ACM New York, NY}
}

@article{lewis2020retrieval,
  title={Retrieval-augmented generation for knowledge-intensive nlp tasks},
  author={Lewis, Patrick and Perez, Ethan and Piktus, Aleksandra and Petroni, Fabio and Karpukhin, Vladimir and Goyal, Naman and K{\"u}ttler, Heinrich and Lewis, Mike and Yih, Wen-tau and Rockt{\"a}schel, Tim and others},
  journal={Advances in neural information processing systems},
  volume={33},
  pages={9459--9474},
  year={2020}
}

@article{chen2021evaluating,
  title={Evaluating large language models trained on code},
  author={Chen, Mark and Tworek, Jerry and Jun, Heewoo and Yuan, Qiming and Pinto, Henrique Ponde De Oliveira and Kaplan, Jared and Edwards, Harri and Burda, Yuri and Joseph, Nicholas and Brockman, Greg and others},
  journal={arXiv preprint arXiv:2107.03374},
  year={2021}
}

@misc{anthropic2025harness,
  title={Effective harnesses for long-running agents},
  author={Young, Justin and Anthropic},
  year={2025},
  howpublished={\url{https://www.anthropic.com/engineering/effective-harnesses-for-long-running-agents}}
}

@misc{anthropic2026harness,
  title={Harness design for long-running application development},
  author={Rajasekaran, Prithvi and Anthropic},
  year={2026},
  howpublished={\url{https://www.anthropic.com/engineering/harness-design-long-running-apps}}
}

@article{arnott2014review,
  title={A critical analysis of decision support systems research revisited: the rise of design science},
  author={Arnott, David and Pervan, Graham},
  journal={Journal of Information Technology},
  volume={29},
  number={4},
  pages={269--293},
  year={2014}
}

@article{wang2024survey,
  title={A survey on large language model based autonomous agents},
  author={Wang, Lei and Ma, Chen and Feng, Xueyang and Zhang, Zeyu and Yang, Hao and Zhang, Jingsen and Chen, Zhiyuan and Tang, Jiakai and Chen, Xu and Lin, Yankai and others},
  journal={Frontiers of Computer Science},
  volume={18},
  number={6},
  pages={186345},
  year={2024},
  publisher={Springer}
}

@article{kraus2020deep,
  title={Deep learning in business analytics and operations research: Models, applications and managerial implications},
  author={Kraus, Mathias and Feuerriegel, Stefan and Oztekin, Asil},
  journal={European Journal of Operational Research},
  volume={281},
  number={3},
  pages={628--641},
  year={2020}
}

@book{wohlin2012experimentation,
  title={Experimentation in software engineering},
  author={Wohlin, Claes and Runeson, Per and H{\"o}st, Martin and Ohlsson, Magnus C and Regnell, Bj{\"o}rn and Wessl{\'e}n, Anders and others},
  volume={236},
  year={2012},
  publisher={Springer}
}

@inproceedings{wang2023planandsolve,
  title={Plan-and-solve prompting: Improving zero-shot chain-of-thought reasoning by large language models},
  author={Wang, Lei and Xu, Wanyu and Lan, Yihuai and Hu, Zhiqiang and Lan, Yunshi and Lee, Roy Ka-Wei and Lim, Ee-Peng},
  booktitle={Proceedings of the 61st annual meeting of the association for computational linguistics (volume 1: long papers)},
  pages={2609--2634},
  year={2023}
}

@article{mosqueira2023human,
  title={Human-in-the-loop machine learning: a state of the art},
  author={Mosqueira-Rey, Eduardo and Hern{\'a}ndez-Pereira, Elena and Alonso-R{\'\i}os, David and Bobes-Bascar{\'a}n, Jos{\'e} and Fern{\'a}ndez-Leal, {\'A}ngel},
  journal={Artificial Intelligence Review},
  volume={56},
  number={4},
  pages={3005--3054},
  year={2023},
  publisher={Springer}
}

@article{jiang2025towards,
  title={Towards Enterprise-Specific Question-Answering for IT Operations and Maintenance based on Retrieval-Augmented Generation Mechanism},
  author={Jiang, Zhuoxuan and Zhang, Tianyang and Bai, Shengguang and Lin, Lin and Zhang, Haotian and Xun, Yinong and Ren, Jiawei and Si, Wen and Zhang, Shaohua},
  journal={Expert Systems with Applications},
  pages={130961},
  year={2025},
  publisher={Elsevier}
}

@article{wang2026intelligent,
  title={Intelligent development of manufacturing enterprises and supply chain Resilience: A perspective based on internal capabilities and external linkages},
  author={Wang, Siwen and Chen, Xiangyi and Han, Na},
  journal={Expert Systems with Applications},
  pages={131338},
  year={2026},
  publisher={Elsevier}
}

@article{chu2026human,
  title={Human--AI Co-Intelligence in smart SMEs: integrating cognitive mechanisms into AI-driven decision-making},
  author={Chu, Kuo-Ming},
  journal={Expert Systems with Applications},
  pages={132373},
  year={2026},
  publisher={Elsevier}
}

@article{sadak2026multi,
  title={A multi-agent LLM framework with Bayesian fusion and safety guardrails for ATC-pilot communication error detection},
  author={Sadak, Mustafa Semih},
  journal={Expert Systems with Applications},
  volume={321},
  pages={132241},
  year={2026},
  publisher={Elsevier}
}

@article{wu2026intention,
  title={Intention-behavior consistency-based automated failure attribution for LLM-driven multi-agent systems},
  author={Wu, Hua and Zheng, Wanhao and Zhang, Minghao and Bai, Xiaojing and Pu, Mengyang and Sun, Li and Liu, Junchen and Xu, Yuhan},
  journal={Expert Systems with Applications},
  pages={133044},
  year={2026},
  publisher={Elsevier}
}

@article{li2026penexpert,
  title={PenExpert: A Multi-Agent Hybrid LLM--Expert System Framework for Autonomous Penetration Testing},
  author={Li, Zhi and Zhou, Tianyang and Zhu, Junhu and Li, Dongyang and Zhu, Xiaodong and Wei, Dongze and Liu, Jinghu and Song, Mengbo},
  journal={Expert Systems with Applications},
  pages={133284},
  year={2026},
  publisher={Elsevier}
}

@article{kwon2001multi,
  title={A multi-agent intelligent system for efficient ERP maintenance},
  author={Kwon, Oh Byung and Lee, JJ},
  journal={Expert Systems with Applications},
  volume={21},
  number={4},
  pages={191--202},
  year={2001},
  publisher={Elsevier}
}

@article{liu2011empirical,
  title={Empirical study on influence of critical success factors on ERP knowledge management on management performance in high-tech industries in Taiwan},
  author={Liu, Pang-Lo},
  journal={Expert Systems with applications},
  volume={38},
  number={8},
  pages={10696--10704},
  year={2011},
  publisher={Elsevier}
}

\end{document}